\documentclass[10pt,twocolumn,letterpaper]{article}

\usepackage{cvpr}              % To produce the CAMERA-READY version
\usepackage[dvipsnames]{xcolor}

\usepackage[pagebackref,breaklinks,colorlinks]{hyperref}

\usepackage{algorithm}
\usepackage{algorithmic} % 为了加算法

\usepackage{booktabs} % 为了表格正常显示
\usepackage{multirow}
\usepackage[figuresright]{rotating}

\usepackage{color} % 为了颜色标签
\usepackage{amssymb}
\usepackage{stmaryrd} % 为了艾弗森括号
\usepackage{makecell} % 为了表内换行

\definecolor{car}{rgb}{0.39215686, 0.58823529, 0.96078431}
\definecolor{bicycle}{rgb}{0.39215686, 0.90196078, 0.96078431}
\definecolor{motorcycle}{rgb}{0.11764706, 0.23529412, 0.58823529}
\definecolor{truck}{rgb}{0.31372549, 0.11764706, 0.70588235}
\definecolor{other-vehicle}{rgb}{0.39215686, 0.31372549, 0.98039216}
\definecolor{person}{rgb}{1.        , 0.11764706, 0.11764706}
\definecolor{bicyclist}{rgb}{1.        , 0.15686275, 0.78431373}
\definecolor{motorcyclist}{rgb}{0.58823529, 0.11764706, 0.35294118}
\definecolor{road}{rgb}{1.        , 0.        , 1.        }
\definecolor{parking}{rgb}{1.        , 0.58823529, 1.        }
\definecolor{sidewalk}{rgb}{0.29411765, 0.        , 0.29411765}
\definecolor{other-ground}{rgb}{0.68627451, 0.        , 0.29411765}
\definecolor{building}{rgb}{1.        , 0.78431373, 0.        }
\definecolor{fence}{rgb}{1.        , 0.47058824, 0.19607843}
\definecolor{vegetation}{rgb}{0.        , 0.68627451, 0.        }
\definecolor{trunk}{rgb}{0.52941176, 0.23529412, 0.        }
\definecolor{terrain}{rgb}{0.58823529, 0.94117647, 0.31372549}
\definecolor{pole}{rgb}{1.        , 0.94117647, 0.58823529}
\definecolor{traffic-sign}{rgb}{1.        , 0.        , 0.    } 

\def\paperID{18} % *** Enter the Paper ID here
\def\confName{3DV\xspace}
\def\confYear{2024\xspace}

\title{PanoSSC: Exploring Monocular Panoptic 3D Scene  Reconstruction for Autonomous Driving}

\author{Yining Shi$^{1,3*}$, Jiusi Li$^{1*}$, Kun Jiang$^{1\dagger}$, Ke Wang$^{2}$, Yunlong Wang$^{1}$, Mengmeng Yang$^{1}$, Diange Yang$^{1\dagger}$   \\
  $^{1}$ School of Vehicle and Mobility, Tsinghua University,
  $^{2}$ Kargobot
  $^{3}$ DiDi Chuxing
}
\begin{document}
\maketitle

\renewcommand{\thefootnote}{\fnsymbol{footnote}}
\footnotetext[2]{This work was done during Yining Shi's internship at DiDi Chuxing. $*$: Yining Shi and Jiusi Li contributed equally to this work.
$\dagger$: Corresponding authors: Kun Jiang, Diange Yang (jiangkun@mail.tsinghua.edu.cn, ydg@mail.tsinghua.edu.cn). 

This work was supported in part by the National Natural Science Foundation of China under Grants (U22A20104, 52372414, 52102464). This work was also sponsored by Tsinghua University-DiDi Joint Research Center for Future Mobility.}

\begin{abstract}
Vision-centric occupancy networks, which represent the surrounding environment with uniform voxels with semantics, have become a new trend for safe driving of camera-only autonomous driving perception systems, as they are able to detect obstacles regardless of their shape and occlusion. Modern occupancy networks mainly focus on reconstructing visible voxels from object surfaces with voxel-wise semantic prediction. Usually, they suffer from inconsistent predictions of one object and mixed predictions for adjacent objects. These confusions may harm the safety of downstream planning modules. To this end, we investigate panoptic segmentation on 3D voxel scenarios and propose an instance-aware occupancy network, PanoSSC. We predict foreground objects and backgrounds separately and merge both in post-processing. For foreground instance grouping, we propose a novel 3D instance mask decoder that can efficiently extract individual objects. we unify geometric reconstruction, 3D semantic segmentation, and 3D instance segmentation into PanoSSC framework and propose new metrics for evaluating panoptic voxels. Extensive experiments show that our method achieves competitive results on SemanticKITTI semantic scene completion benchmark.

% Yining: Put this paragraph to methodology:overview
% Our method uses a tri-perspective view encoder\cite{TPVFormer} to lift 2D image features to 3D space. Along with a lightweight 3D semantic occupancy prediction head, these 3D features are passed through a novel 3D instance mask decoder to improve completion performance of the foreground. 
 
\end{abstract}
\section{Introduction}\label{sec:intro}
\begin{figure}[htbp]
	\centering
	\includegraphics[width=0.47\textwidth]{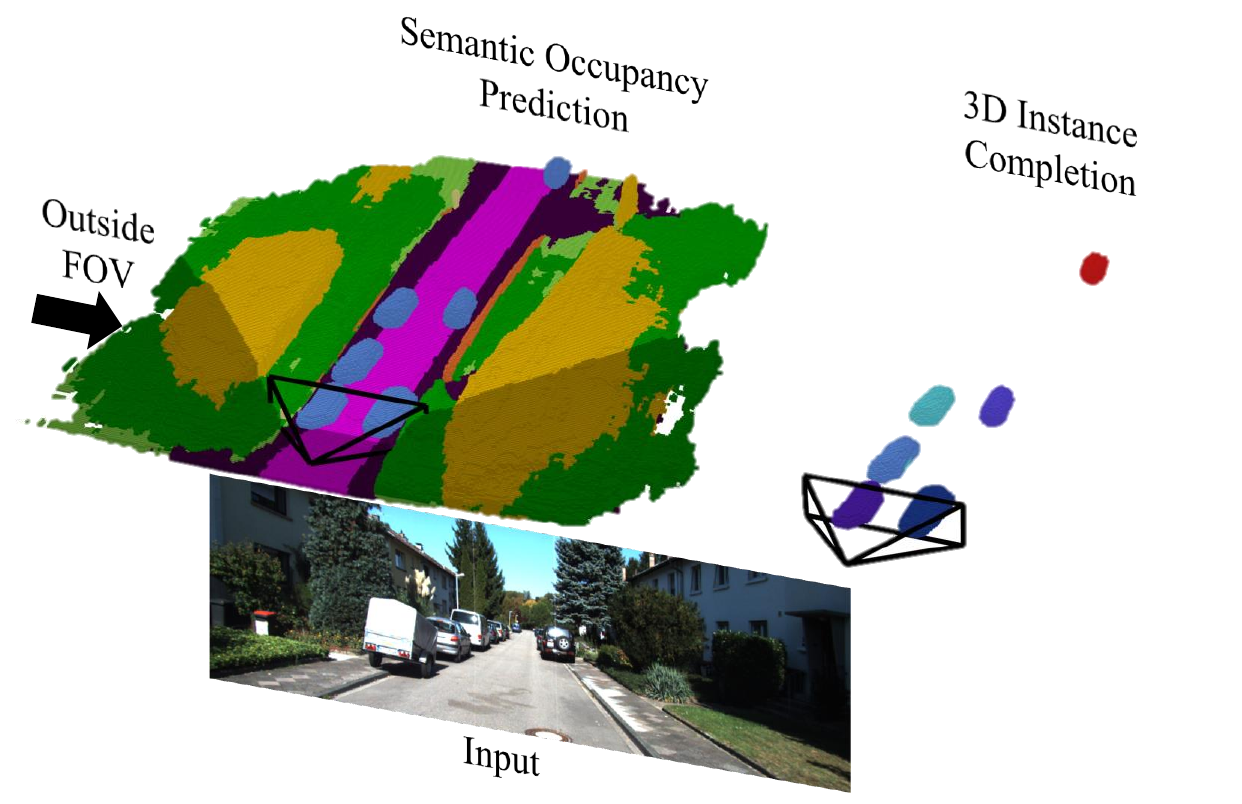}
	\caption{Panoptic 3D scene reconstruction from a monocular RGB image for outdoor scenes with PanoSSC. Our method infers voxel-level occupancy, semantics and instance ids.
	}
	\label{fig:intro}
\end{figure}

Accurate understanding of the 3D surroundings is an essential prerequisite for safe autonomous driving systems. Apart from the mature object-centric perception pipelines which consist of detection, tracking and prediction \cite{3DODSurvey}, the newly emerging occupancy networks cast new insights for fine-grained scene understanding \cite{Occ3D,GridCentricSurvey}. Occupancy networks are more capable of representing partly occluded, deformable, or semantically not well-defined obstacles and conducting open-world general object detection, so recently they are widely investigated in both academia and industry.

Reconstructing the surroundings as 3D voxels originates from semantic scene completion (SSC) from a single LiDAR frame. Since Tesla announced its vision-only occupancy network, various vision-centric occupancy networks \cite{TPVFormer, OpenOccupancy, OccNet} are proposed with additional voxel-level labels and occupancy prediction benchmarks on nuScenes, KITTI360, and Waymo open dataset \cite{SSCBench, Occ3D, OpenOccupancy, OccNet}. 
%Occupancy prediction focuses on voxel regions visible to cameras, mainly the nearest object surface to the ego vehicle. 

% 这种数据集的缺陷只在semantickitti里
% Regarding different benchmarks, SSC focuses on entire occupancy state of most static objects, while dynamic objects yield ghost trails.

Although recent vision-based methods perform as well as LiDAR-based methods on segmentation task \cite{TPVFormer,OccNet}, \textbf{instances extraction} in semantic mapping are less explored. Understanding instances in the environment is able to eliminate inconsistent semantic predictions of one object and mixed predictions for adjacent objects, while these confusions may harm the safety of downstream planning modules. We intend to conduct instance-aware semantic occupancy prediction on SSC benchmarks since SSC tasks require an entire representation of a single object. A concurrent work, PanoOcc \cite{PanoOcc, UniOcc}, conducts panoptic segmentation on LiDAR panoptic benchmark via multi-task learning of occupancy prediction and object detection. The main difference between this paper and PanoOcc \cite{PanoOcc} lies in that we don't assume labeling objects as bounding boxes and only learn instances with segmentation labels. Hence, we are motivated to adapt to diverse environments with obstacles in which bounding boxes do not fit.

% MonoScene\cite{MonoScene} is the pioneering vision-based work, which infers the dense semantic occupancy scene from a single image. However, it is inefficient to obtain voxel-based features by projection and then process them with 3D UNet. In addition, the SSC methods only predict semantics of voxels without further distinguishing different object instances.

We propose PanoSSC, a novel monocular panoptic 3D scene reconstruction method. PanoSSC consists of image encoder, 2D to 3D transformer, semantic occupancy head and transformer-based mask decoder head. Image features are lifted to 3D space for 3D semantic occupancy prediction and 3D instance completion. Unlike previous semantic occupancy networks that adopt a per-voxel classification formulation, we design a 3D mask decoder for foreground instance completion and perform mask-wise classification. This design is motivated by an insight: semantic segmentation and instance segmentation for 2D images can benefit from multi-task learning \cite{UncertaintyWeighLosses,PanopticFPN}. Similar to \cite{PanopticSegformer}, we propose a strategy for merging results of the two heads to obtain voxel-level occupancy, semantics and instance ids. An graphical illustration is shown in \cref{fig:intro}. 
% Besides, we apply indoor panoptic 3D scene reconstruction\cite{panoptic3dss} to large-scale traffic scenarios and evaluate our method on this task. 

%\IEEEpubidadjcol % 此命令必须在第一页右栏的文字中调用
In summary, our main contributions are listed as follows:
\begin{enumerate}
	\item We propose the task of panoptic 3D scene reconstruction for outdoor scenes, aiming to predict voxel-level occupancy, semantics and instance id.
	\item We propose a novel monocular semantic occupancy network, PanoSSC, which includes two prediction heads to perform semantic occupancy prediction and 3D instance completion respectively. The joint learning of these two heads can promote each other.
	\item  Our method achieves competitive semantic occupancy prediction results compared to the monocular pioneering work on SemanticKITTI \cite{SemanticKITTI}. It is also the first to tackle panoptic 3D semantic scene reconstruction on outdoor.
\end{enumerate}
\section{Related works}\label{sec:related_works}

% \subsection{Vision-centric perception}
% The overall pipeline of vision-centric perception:  we need to perceive what , data-driven view transformation, head design

% \textbf{Semantic scene completion.}
\textbf{Semantic occupancy prediction.}
% Semantic scene completion, as the initial voxelized scene understanding technique, 
Semantic occupancy prediction, originally called semantic scene completion (SSC), is introduced in SSCNet \cite{SSCNet} for indoor scenes, which aims to jointly address 3D semantic segmentation and 3D scene completion and achieve mutual promotion. Since then, many SSC methods on indoor have been proposed, which directly use depth images \cite{AICNet,DDR} from RGB-D as input or encode depth information as occupancy grids \cite{TS3D,SCFusion} or TSDF \cite{SSCNet,SGC,3DSketch}. SemanticKITTI \cite{SemanticKITTI} is the first large-scale dataset that proposes this task for LiDAR in the real outdoor world.
%, which greatly promotes the development of voxelized scene understanding for autonomous driving. 
Most methods \cite{LMSCNet,S3CNet,JS3CNet,MotionSC,Local-DIFs} for outdoor depend on LiDAR point clouds.
%and use a UNet architecture with 2D or 3D CNNs to obtain long-range contextual information. 
After Tesla's Occupancy Network, semantic occupancy prediction based on low-cost cameras has received extensive attention.
% and the SSC task is also called semantic occupancy prediction. 
MonoScene \cite{MonoScene} is the first to infer dense 3D voxelized semantic scenes from a single RGB image.
%, which proposes a projection and sampling mechanism to bridge successive 2D and 3D UNet. 
%OccDepth \cite{OccDepth} further considers the correlation  between stereo images and uses implicit depth information to help the reconstruction of 3D structures.
OccDepth \cite{OccDepth} further uses implicit depth information from stereo images for 3D structure reconstruction.
To avoid the ambiguity of 3D features caused by occlusion, VoxFormer \cite{Voxformer} first forms sparse 3D voxel features of the visible area and then densifies them. TPVFormer \cite{TPVFormer} proposes an efficient tri-perspective view representation to replace voxel-based features and generates occupancy prediction with multi-view images. Our method is able to perform semantic occupancy prediction from a monocular image, and further distinguish different instances belonging to the same foreground category.

\textbf{Semantic and panoptic segmentation.}
Semantic and panoptic segmentation are thoroughly investigated with the development of deep learning. Since FCNs \cite{FCNs}, semantic segmentation mainly relies on per-pixel classification, while mask classification dominates for instance-level segmentation tasks \cite{instseg,panop}.
%Semantic segmentation mainly relies on convolution decoder on per-pixel feature maps\cite{FCNs}, while panoptic segmentation methods mostly benefit from mask decoders\cite{instseg,panop}.  
In 2D domain, early mask-based methods \cite{MaskR-CNN,DETR} first predict bounding boxes and then generate a binary mask for each box, while others \cite{MaskFormer,Max-DeepLab,PanopticSegformer} discard the boxes and directly predict masks and categories. In the automotive perception domain, vision bird's eye view (BEV) algorithms \cite{LSS,PETRv2,BEVFormer} focus on the transformation from perspective view (PV) to BEV and segment drivable areas, lanes and vehicles on BEV. 
%And BEV feature maps can easily be extended to different segmentation tasks. 
3D panoptic segmentation are designed for sparse LiDAR point clouds \cite{LiDARMultiNet}. Dahnert et al. \cite{panoptic3dss} unify the tasks of geometric reconstruction, 3D semantic segmentation, and 3D instance segmentation into panoptic 3D scene reconstruction for indoor scenes, and propose a monocular method. PNF \cite{PNF} generates panoptic neural scene representation with self-supervision from an RGB sequence, while it focuses on offline reconstruction like most other NeRF-style methods rather than real-time semantic occupancy prediction. We address outdoor panoptic 3D scene reconstruction and generate 3D voxel binary mask for each object to conduct mask classification and instance-aware semantic occupancy prediction.

\textbf{Multi-task learning.} 
%Multi-task learning aims to improve learning efficiency and prediction accuracy by sharing domain information between related tasks. 
For images, many works \cite{UncertaintyWeighLosses,PanopticFPN} regard semantic segmentation and instance segmentation as related tasks for joint learning and achieve good performance. For autonomous driving, multi-task learning is widely used in LiDAR semantic segmentation. LidarMultiNet \cite{LiDARMultiNet} is a unified framework  for 3D semantic segmentation, object detection and panoptic segmentation. JS3CNet \cite{JS3CNet} exploits the shape priors from semantic scene completion to improve the performance of segmentation. Inspired by these works, we design two heads for semantic occupancy prediction and 3D instance completion respectively, and conduct joint learning to achieve mutual promotion.
% Yining: merging panoptic and multi-task learning into semantic segmentation is enough 

\begin{figure*}[ht]
	\centering
	\includegraphics[width=\textwidth]{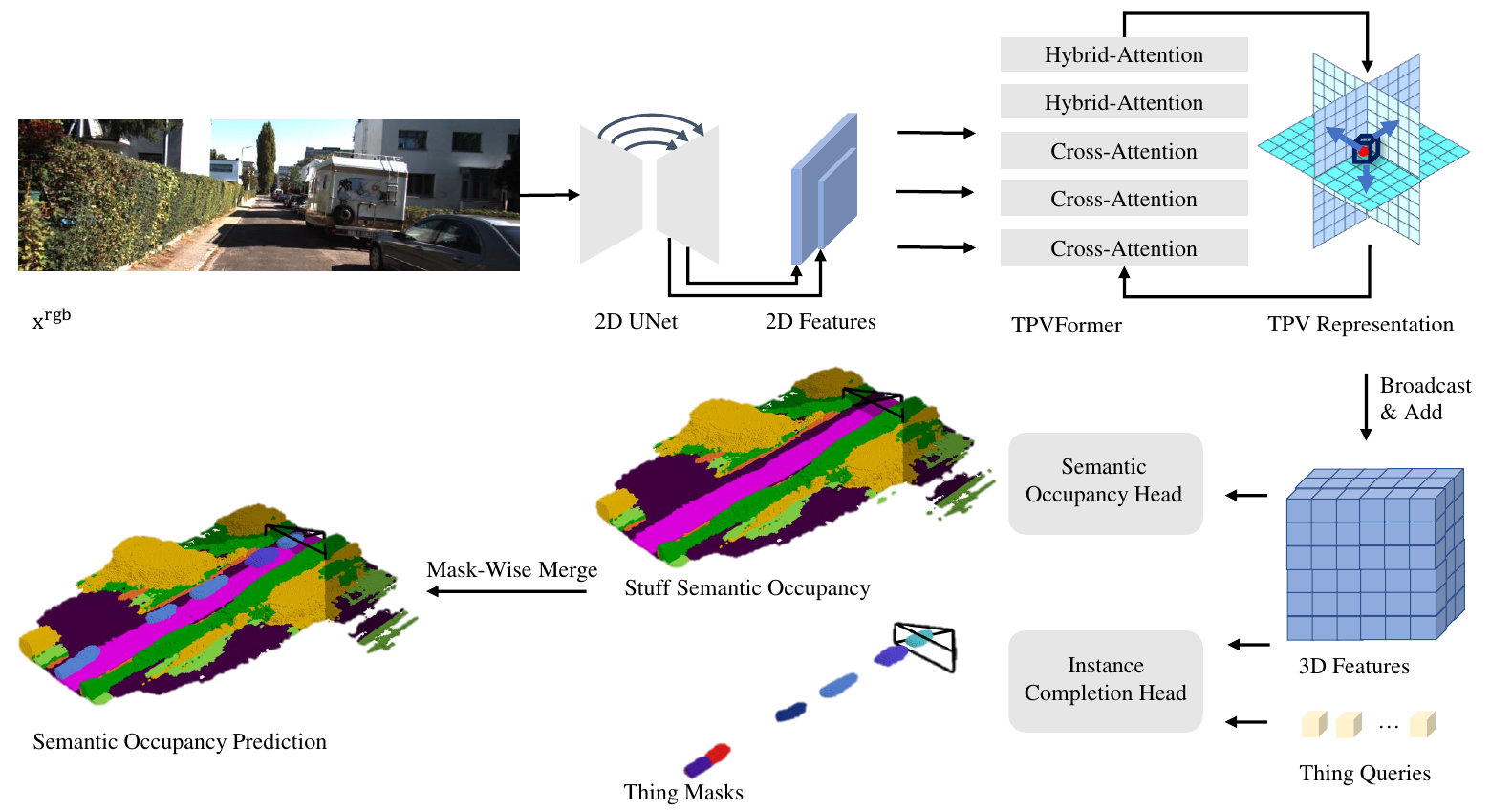}
	%\vspace{-7mm}
	\caption{PanoSSC framework. We adopt 2D UNet to generate multi-scale image features and lift them to 3D space with TPVFormer \cite{TPVFormer}. After broadcasting TPV features, the voxel features are used for 3D semantic occupancy prediction and instance completion respectively. During inference, we adopt a mask-wise strategy to merge the results of two prediction heads.
	}
	\label{fig:occ_framework}
	%\vspace{-6mm}
\end{figure*}
\section{Methodology}\label{sec:methodology}

\subsection{Architecture}
Semantic occupancy prediction is to discretize 3D scene into voxels and assign each voxel a semantic label $C=\left \{c_{0},c_{1},...,c_{N}\right \}$, where $c_{0}$ denotes free class and $N$ is the number of interested semantic classes. Similar to \cite{panoptic3dss}, panoptic 3D scene reconstruction is to further predict instance id for each voxel belonging to foreground categories.

Our architecture, PanoSSC, shown in \cref{fig:occ_framework}, solves the above tasks given only a single RGB image. The architecture starts from an arbitrary image backbone, and then a view transformation module, TPVFormer \cite{TPVFormer} in our implementation, to lift image features to 3D space. After that, we broadcast each TPV feature along the orthogonal direction and add them to obtain voxel feature. Along with a lightweight MLP-based semantic occupancy head, these voxel features are passed through a novel 3D mask decoder (\cref{sec:maskdecoder}) to improve the completion performance of the foreground instances. Under our training strategy (\cref{sec:train}), these two prediction heads are able to boost each other. Inspired by Panoptic SegFormer \cite{PanopticSegformer}, we employ a mask-wise strategy (\cref{sec:merge}) to merge predicted 3D masks from the final mask decoder layer with the background results from semantic occupancy head to obtain occupancy, semantics and instance ids for 3D voxelized scene.

\textbf{2D-3D encoder.} For fair comparisons with monocular pioneering work \cite{MonoScene} on semantic occupancy prediction task, we employ the 2D UNet based on the pretrained EfficientNet-B7 \cite{EfficientNet} to generate multi-scale feature maps, of which the resolutions are $1/8$, $1/16$ compared to the input image. Then we use linear layers to convert the feature dimension to $96$ and send them to TPVFormer \cite{TPVFormer}.  We follow the settings in \cite{TPVFormer} to stack 3 hybrid-cross-attention block (HCAB) blocks and 2 hybrid-attention block (HAB) blocks to form TPVFormer and set the number of queries on TPV planes to be $128\times128$, $16\times128$, $128\times16$. Each query encodes features of pillar region above the grid cell belonging to one of the TPV planes.

% 这一段写supl material里面，介绍别人的意义不大
% TPVFormer is a transformer-based 3D encoder and proposes two kinds of deformable attention: image cross-attention (ICA) for lifting image features to TPV planes and cross-view hybrid-attention (CVHA) for interactions among TPV planes and contextual information encoding. Hybrid-cross-attention block (HCAB) consists of CVHA and ICA and hybrid-attention block (HAB) contains only CVHA.

\textbf{Semantic occupancy head.} To obtain full-scale voxel features of size $H\times W\times D\times C$ for fine-grained segmentation, we perform bilinear interpolation on the TPV features, and then broadcast each plane along the orthogonal direction and add them together. After that, the voxel features are fed into an MLP-based semantic occupancy head to predict their semantic labels, which consists of only two linear layers and an intermediate activation layer.

\begin{figure}[ht]
	\centering
	\includegraphics[width=0.47\textwidth]{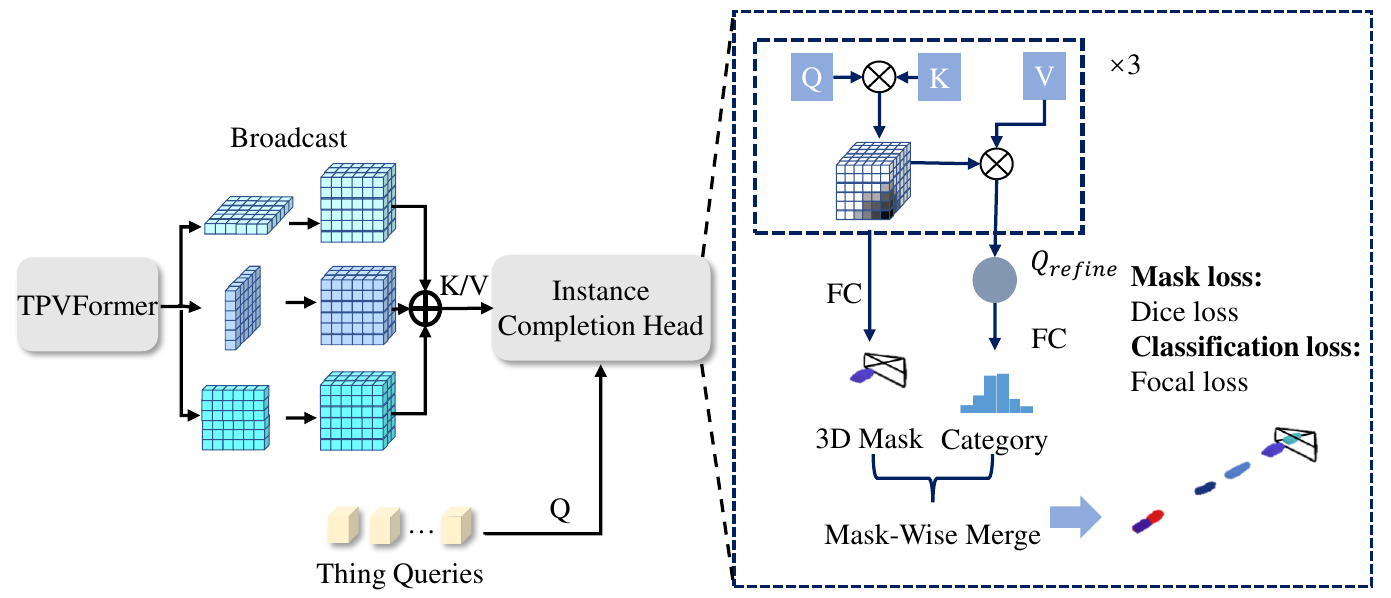}
	%\vspace{-7mm}
	\caption{3D mask decoder. We input the voxel features from TPVFormer \cite{TPVFormer} and the initialized thing queries into the transformer-based 3D mask decoder, which can generate 3D instance masks from attention maps and probabilities over all foreground categories from refined queries.
	}
	\label{fig:occ_maskdecoder}
	%\vspace{-6mm}
\end{figure}

\subsection{3D mask decoder}\label{sec:maskdecoder}
To improve the reconstruction and segmentation quality of foreground instances, we also feed the voxel features into an instance completion head to conduct instance-aware semantic occupancy prediction. We propose a transformer-based 3D mask decoder as the instance completion head to predict categories and 3D masks from given queries, as shown in \cref{fig:occ_maskdecoder}. We initialize $N$ learnable reference points with uniform distribution from 0 to 1 in 3D space and involve positional encoding \cite{Attention}. Then they are passed through an MLP consisting of two linear layers to generate the initial thing queries $Q$. The keys $K$ and values $V$ are projected from the voxel features. Specifically, due to computational cost constraints, TPV features are down-sampled by two convolution layers and an average pooling layer, and then broadcast to obtain 3D voxel features of size $H/4 \times W/4 \times D/4 \times 256(C)$. 

3D mask decoder is stacked by multiple transformer layers, and each layer can generate attention maps $A\in \mathbb{R}^{N\times h \times (\frac{H}{4}\times \frac{W}{4}\times \frac{D}{4})}$ and refined queries $Q_{refined}\in \mathbb{R}^{N\times 256}$,where $h$ is the number of attention heads. This process can be formulated as:
\begin{align}
	A &=\frac{QK^{T}}{\sqrt{d_{k}}}\\
	Q_{refined} &= \mathrm{softmax}(A)\cdot V
\end{align}
where $d_{k}$ is the dimension of $Q$ and $K$. We use $N=300$, $h=8$ and stack 3 layers.

For $Q_{refined}$ from each layer, a FC layer is used to directly predict probabilities over all foreground categories. At the same time, we use a linear layer to fuse the attention maps of multiple attention heads to obtain 3D masks $M \in \mathbb{R}^{N\times (\frac{H}{4}\times \frac{W}{4}\times \frac{D}{4})}$.

% 这段意义不大，可以放supplimentary里面
% We observe that using a lightweight FC layer to generate masks from attention maps enables the attention module to learn where to focus guided by the ground truth mask. We extend the mask decoder in \cite{PanopticSegformer} to 3D segmentation and completion, and also use deep supervision, which means that attention maps of each layer are supervised by the ground truth 3D mask. Therefore, the attention module can focus on interested region as early as possible, which accelerating the learning and convergence of the model. 

\subsection{Training strategy}\label{sec:train}
PanoSSC includes the multi-task learning from both semantic occupancy head and instance completion head. In multi-task learning, a common approach is to perform a weighted linear sum of the losses for each task \cite{UncertaintyWeighLosses}. But model performance heavily relies on weight selection.
%, and simply adding the losses is likely to degrade the performance of one of the tasks. 
To get better results, we train our network in a fine-tuning two-stage manner.

At the first stage, we only train the network without instance completion head, which only consists of 2D UNet, TPVFormer and semantic occupancy head. We consider the semantic occupancy prediction task as a pre-training step. At this stage, in addition to the most commonly used weighted cross-entropy loss $\mathcal{L}_{\text{ce}}$ for semantic occupancy prediction, we also use the scene-class affinity loss $\mathcal{L}^{\text{sem}}_{\text{scal}}$, $\mathcal{L}^{\text{geo}}_{\text{scal}}$ and frustum proportion loss $\mathcal{L}_{\text{fp}}$ proposed in \cite{MonoScene} to optimize the global and local performance on this task. So the loss function at the first stage writes:
\begin{equation}
	\mathcal{L}_\text{seg} = \mathcal{L}_{\text{ce}}
	+ \mathcal{L}^{\text{sem}}_{\text{scal}}
	+ \mathcal{L}^{\text{geo}}_{\text{scal}}
	+ \mathcal{L}_{\text{fp}}.
\end{equation}

Instance completion head can generate a fixed-size prediction set.
%, and each prediction includes the category probability and the 3D mask. 
Similar to transformer-based works \cite{DETR,PanopticSegformer}, we use Hungarian algorithm \cite{Hungarian} to obtain the best bipartite matching between the prediction set and the ground truth set.
%with the minimum matching cost. 
The matching cost is the sum of classification cost and mask cost (classification loss $\mathcal{L}_\text{cls}$ and mask loss $\mathcal{L}_\text{mask}$). The loss for instance completion head is defined as:
\begin{equation}
	\mathcal{L}_\text{inst} =\sum_{i}^{D_m}{(
		\lambda_\text{cls}\mathcal{L}_\text{cls}^{i} + \lambda_\text{mask}\mathcal{L}_\text{mask}^{i})},
  \label{eq:maskloss}
\end{equation}
where $D_m$ is the number of layers in 3D mask decoder, $\lambda_\text{cls}$ and $\lambda_\text{mask}$ are the weights. We employ focal loss \cite{focalloss} as classification loss $\mathcal{L}_\text{cls}$ and dice loss \cite{diceloss} as mask loss $\mathcal{L}_\text{mask}$. In practice, we use $D_m=3$, $\lambda_\text{cls}=1$, $\lambda_\text{mask}=2$.

At the second stage, we add instance completion head and reduce the learning rate of the rest of the network for joint training. The loss function at the second stage writes:
\begin{equation}
	\mathcal{L} = \mathcal{L}_{\text{seg}}
	+ \mathcal{L}_{\text{inst}}.
\end{equation}

\subsection{Mask-wise merging inference}\label{sec:merge}
This stage further refines the reconstruction quality of the foreground instances. We design a mask-wise merging strategy for 3D masks. During inference, it only takes the background prediction results of semantic occupancy head, and sets the voxels which belong to the foreground categories to empty. Then 3D masks from the instance completion head are merged one by one into the semantic occupancy prediction result. Since each mask represents a foreground instance, a unique id can be assigned. So PanoSSC can address the panoptic 3D scene reconstruction task. 

Similar to \cite{PanopticSegformer}, we calculates the confidence scores of 3D masks to determine the category and id of the overlap region. These scores consist of classification probabilities and mask quality scores. The score of i-th prediction writes:
\begin{equation}
	s_{i}=p_{i}^{\alpha } \times \left ( \frac{\sum m_{i}[h,w,d]\llbracket m_{i}[h,w,d]>0.25\rrbracket}{\sum \llbracket m_{i}[h,w,d]>0.25\rrbracket}  \right )^{\beta },
\end{equation}
where $\llbracket.\rrbracket$ is the Iverson bracket, $p_{i}$ is the maximum classification probability of i-th result, $m_{i}[h,w,d]$ is the mask logit at voxel $[h,w,d]$, $\alpha,\beta$ are employed to balance the weight of classification probability and mask quality. In practice, we use $\alpha =\frac{1}{3}$, $\beta =1$. Note that since the resolution of 3D masks generated by the instance completion head is $H/4 \times W/4 \times D/4$, we perform trilinear interpolation to obtain full-scale masks, and then the masks are binarized with a threshold of 0.25.

\cref{alg:maskmerge} illustrates our mask-wise merging strategy. It takes predicted categories $c$, confidence scores $s$ and 3D masks $m$ as input. These prediction results are arranged in descending order of confidence scores. In addition, the field of view of the image $FOV$ is also input, in which the inside voxels are 1 and the outside voxels are 0. We set all voxels belonging to the foreground categories in the result from semantic occupancy head to 0 as the initial value of $SemResult$. And instance id result $IdResult$ is initialized by zeros. 
%This strategy outputs $SemResult$ and $IdResult$ to assign a category label and an instance id to each foreground voxel. 

We merge masks into the final result in order and discard all masks with confidence scores below $t_{q}$. Then, we take the intersection of the current binarized mask and the empty voxels in $SemResult$ to obtain non-overlap part $m_{i}$ of the mask. If the proportion of $m_{i}$ to the origin mask is lower than $t_{overlap}$, it is considered that there is a overlap conflict and the mask need to be discarded. Due to the extremely low prediction accuracy of instances outside the FOV, only masks that are mostly within $FOV$ (above $t_{fov}$) will be kept. Finally, the category label and instance id of each mask are assigned to $SemResult$ and $IdResult$ for panoptic 3D scene reconstruction. In practice, we use $t_{q}=0.2$, $t_{overlap}=0.5$, $t_{fov}=0.5$.

\begin{algorithm}[t]
	\caption{Mask-Wise Merging.}\label{alg:maskmerge}
	\renewcommand{\algorithmicrequire}{\textbf{Input:}}
	\renewcommand{\algorithmicensure}{\textbf{Output:}}
	\begin{algorithmic}[1]
		\REQUIRE background semantic result $SemResult \in \mathbb{R}^{H\times W\times D}$, the field of view of the image $FOV \in \mathbb{R}^{H\times W\times D}$, instance id result $IdResult \in \mathbb{R}^{H\times W\times D}$, categories $c \in \mathbb{R}^{N}$, scores $s \in \mathbb{R}^{N}$, masks $m \in \mathbb{R}^{N\times H\times W\times D}$.
		\ENSURE semantic result $SemResult$, instance id result $IdResult$.
		\STATE Initialize: $IdResult \gets 0, id \gets 1$
		\STATE Sort results in descending order of score: $order$
		\FOR{$i$ in $order$}
		\IF{$s[i]>t_{q}$}
		\STATE $m_{i} \gets (m[i]>0.25) \&  (SemResult=0)$ 
		\IF{$\frac{m_{i}}{m[i]>0.25}>t_{overlap}$ \AND $\frac{m_{i}\& FOV}{m[i]>0.25}>t_{fov}$}
		\STATE $SemResult[m_{i}] \gets c[i]$
		\STATE $IdResult[m_{i}] \gets id$
		\STATE $id \gets id+1$
		\ENDIF
		\ENDIF
		\ENDFOR
	\end{algorithmic}
\end{algorithm}

\section{Experiments}\label{sec:exps}

We evaluate PanoSSC on the densely annotated autonomous driving dataset SemanticKITTI \cite{SemanticKITTI}. In addition to the SSC task, we propose the outdoor panoptic 3D scene reconstruction task and corresponding metrics (\cref{sec:metrics}) based on this dataset. We provide our performance on two tasks (\cref{sec:performance}) and conduct ablation studies (\cref{sec:ablation}).

\subsection{Experimental setup}

\begin{table*}[htbp]
        \footnotesize
        \setlength{\tabcolsep}{0.004\linewidth}
        \newcommand{\classfreq}[1]{{~\tiny(\semkitfreq{#1}\%)}}
        \centering

	\begin{tabular}{l|c|c|ccccccccccccccccccc}
		\toprule
		Method & \begin{sideways}IoU (\%)\end{sideways} & \begin{sideways}mIoU (\%)\end{sideways} & \begin{sideways}Road (15.30\%)\end{sideways} & \begin{sideways}Parking (1.12\%)\end{sideways} & \begin{sideways}Sidewalk (11.13\%)\end{sideways} & \begin{sideways}Other-ground (0.56\%)\end{sideways} & \begin{sideways}Building (14.10\%)\end{sideways} & \begin{sideways}Fence (3.90\%)\end{sideways} & \begin{sideways}Vegetation (39.30\%)\end{sideways} & \begin{sideways}Terrain (9.17\%)\end{sideways} & \begin{sideways}Car (3.92\%)\end{sideways} & \begin{sideways}Bicycle (0.03\%)\end{sideways} & \begin{sideways}Motorcycle (0.03\%)\end{sideways} & \begin{sideways}Truck (0.16\%)\end{sideways} & \begin{sideways}Other-vehicle (0.20\%)\end{sideways} & \begin{sideways}Person (0.07\%)\end{sideways} & \begin{sideways}Bicyclist (0.07\%)\end{sideways} & \begin{sideways}Motorcyclist (0.05\%)\end{sideways} & \begin{sideways}Pole (0.29\%)\end{sideways} & \begin{sideways}Traffic-sign (0.08\%)\end{sideways} & \begin{sideways}Trunk (0.51\%)\end{sideways} \\
		\midrule
		LMSCNet$^{rgb}$ \cite{LMSCNet}* & 28.61 & 6.70  & 40.68 & 4.38  & 18.22 & 0.00  & 10.31 & 1.21  & 13.66 & 20.54 & 18.33 & 0.00  & 0.00  & 0.00  & 0.00  & 0.00  & 0.00  & 0.00  & 0.00  & 0.00  & 0.02 \\
		3DSketch$^{rgb}$ \cite{3DSketch}* & 33.30 & 7.50  & 41.32 & 0.00  & 21.63 & 0.00  & \underline{14.81} & 0.73  & \textbf{19.09} & 26.40 & 18.59 & 0.00  & 0.00  & 0.00  & 0.00  & 0.00  & 0.00  & 0.00  & 0.00  & 0.00  & 0.00 \\
		AICNet$^{rgb}$ \cite{AICNet}* & 29.59 & 8.31  & 43.55 & 11.97 & 20.55 & 0.07  & 12.94 & 2.52  & 15.37 & \underline{28.71} & 14.71 & 0.00  & 0.00  & 4.53  & 0.00  & 0.00  & 0.00  & 0.00  & 0.06  & 0.00  & \underline{2.90} \\
		JS3CNet$^{rgb}$ \cite{JS3CNet}* & \textbf{38.98} & 10.31 & 50.49 & 11.94 & 23.74 & 0.07  & \textbf{15.03} & 3.94  & \underline{18.11} & 26.86 & \textbf{24.65} & 0.00  & 0.00  & 4.41  & \underline{6.15}  & 0.67  & \underline{0.27}  & 0.00  & \underline{3.77}  & \underline{1.45}  & \textbf{4.33} \\
		MonoScene \cite{MonoScene}** & \underline{36.87} & \textbf{11.27} & \underline{55.92} & \underline{14.55} & \textbf{26.51} & \textbf{1.55} & 13.47 & \textbf{6.66} & 17.98 & \textbf{29.90} & \underline{23.34} & \underline{0.24}  & \textbf{0.74} & \underline{9.05}  & 2.59  & \textbf{1.96} & \textbf{1.08} & 0.00  & \textbf{3.84} & \textbf{2.40} & 2.41 \\
        \midrule
		PanoSSC (ours) & 34.94 & \underline{11.22} & \textbf{56.36} & \textbf{17.76} & \underline{26.40} & \underline{0.88}  & 14.26 & \underline{5.72}  & 16.69 & 28.05 & 19.63 & \textbf{0.63} & \underline{0.36}  & \textbf{14.79} & \textbf{6.22} & \underline{0.87}  & 0.00  & 0.00  & 1.94  & 0.70  & 1.83 \\
		\bottomrule
	\end{tabular}%
 	\caption{Semantic scene completion results on SemanticKITTI validation set. (* represents that the results are reported on \cite{MonoScene}. ** represents the reproduced result using the official code and checkpoint.)}
	\label{tab:ssc_result}%
\end{table*}%

\begin{table*}[htbp]
        \footnotesize
	\centering
	\renewcommand\arraystretch{1.0}

	\begin{tabular}{l|rrr|rrr|rrr}
		\toprule
		& \multicolumn{1}{l}{PRQ} & \multicolumn{1}{l}{RSQ} & \multicolumn{1}{l|}{RRQ} & \multicolumn{1}{l}{PRQ} & \multicolumn{1}{l}{RSQ} & \multicolumn{1}{l|}{RRQ} & \multicolumn{1}{l}{PRQ} & \multicolumn{1}{l}{RSQ} & \multicolumn{1}{l}{RRQ} \\
		& \multicolumn{3}{c|}{} & \multicolumn{3}{c|}{things} & \multicolumn{3}{c}{stuff} \\
		\midrule
		MonoScene \cite{MonoScene} + EC & 19.33 & 37.83 & 38.26 & 6.51  & 30.90 & 18.15 & 57.79 & 58.63 & \textbf{98.58} \\
		TPVFormer \cite{TPVFormer} + EC & 18.94 & 32.18 & 36.40 & 6.39  & 23.50 & 16.12 & 56.59 & 58.20 & 97.22 \\
		PanoSSC (ours) & \textbf{22.93} & \textbf{39.51} & \textbf{49.43} & \textbf{11.27} & \textbf{33.02} & \textbf{33.20} & \textbf{57.90} & \textbf{59.00} & 98.13 \\
		\bottomrule
	\end{tabular}%
 	\caption{Panoptic 3D scene reconstruction results on SemanticKITTI validation set. (EC: Euclidean clustering.) }
	\label{tab:panop3dss_result}%
\end{table*}%

\begin{table}[htbp]
        \footnotesize
	\centering
	\setlength\tabcolsep{2.2pt}
	\renewcommand\arraystretch{1.0}

	\begin{tabular}{ccccc}
		\toprule
		&       & \makecell[c]{MonoScene \cite{MonoScene}\\+EC} & \makecell[c]{TPVFormer \cite{TPVFormer}\\+EC} & PanoSSC (ours) \\
		\midrule
		\multirow{3}[2]{*}{Car} & PRQ   & 14.73 & \textbf{16.95} & 16.38 \\
		& RSQ   & 41.03 & \textbf{41.68} & 35.65 \\
		& RRQ   & 35.90 & 40.68 & \textbf{45.95} \\
		\midrule
		\multirow{3}[2]{*}{Truck} & PRQ   & 4.06  & 0.00  & \textbf{13.26} \\
		& RSQ   & 25.89 & 0.00  & \textbf{33.34} \\
		& RRQ   & 15.67 & 0.00  & \textbf{39.78} \\
		\midrule
		\multirow{3}[2]{*}{Other-vehicle} & PRQ   & 0.74  & 2.21  & \textbf{4.17} \\
		& RSQ   & 25.77 & 28.82 & \textbf{30.07} \\
		& RRQ   & 2.88  & 7.68  & \textbf{13.87} \\
		\midrule
		\multirow{3}[2]{*}{Road} & PRQ   & 57.79 & 56.59 & \textbf{57.90} \\
		& RSQ   & 58.63 & 58.20 & \textbf{59.00} \\
		& RRQ   & 98.58 & 97.22 & \textbf{98.13} \\
		\bottomrule
	\end{tabular}%
 	\caption{Panoptic 3D scene reconstruction results for each category on SemanticKITTI validation set. (EC: Euclidean clustering.)}
	\label{tab:panop3dss_result_c}%
\end{table}%

\textbf{Dataset.} The SSC task of SemanticKITTI \cite{SemanticKITTI} focuses on the volume of 51.2m ahead of the car, 25.6m to each side and 6.4m in height, and discretize it into $256\times 256\times 32$ voxels. The voxels are labelled with 21 classes (19 semantics, 1 free and 1 unknown). Similar to previous work \cite{MonoScene}, we left crop RGB images of cam2 to $1220\times 370$. We use the official 3834/815 train/val splits. 
% When training the instance completion head, the 3D binary mask and the category label for each instance is required. While SemanticKITTI only provides ground-truth semantic labels without instance ids. 
To train and evaluate our network, we perform Euclidean clustering on the ground truth of train set and validation set to distinguish different instances. Notice that the dense semantic labels are obtained by the rigid registration of continuous frames \cite{SSCSurvey}, so moving objects (\eg moving people) inevitably produce traces, which is an imperfection of SemanticKITTI. We filter out these traces when clustering. As shown in the supplementary material , by setting the clustering parameters reasonably, we can obtain unique ids for different instances.

\textbf{Training setup.} As mentioned in \cref{sec:train}, we train our network in a two-stage manner. We first jointly pretrain 2D UNet, TPVFormer and semantic occupancy head on 4 RTX 3090 GPUs with an AdamW \cite{AdamW} optimizer using a batch size of 4, learning rate $2\mathrm{e}{\text{-}4}$ and a weight decay of 0.01 for 10 epochs. At the second stage, instance completion head is joined for joint training for another 10 epochs. With other settings unchanged, the learning rate is $1\mathrm{e}{\text{-}4}$ for instance completion head and $1\mathrm{e}{\text{-}5}$ for other parts.

\begin{figure*}
	\centering
	\scriptsize
	\begin{tabular}{cccc}
		Input & MonoScene \cite{MonoScene} & PanoSSC (ours) & Ground Truth\\
		\multirow{2}{*}{\includegraphics[width=0.4\columnwidth]{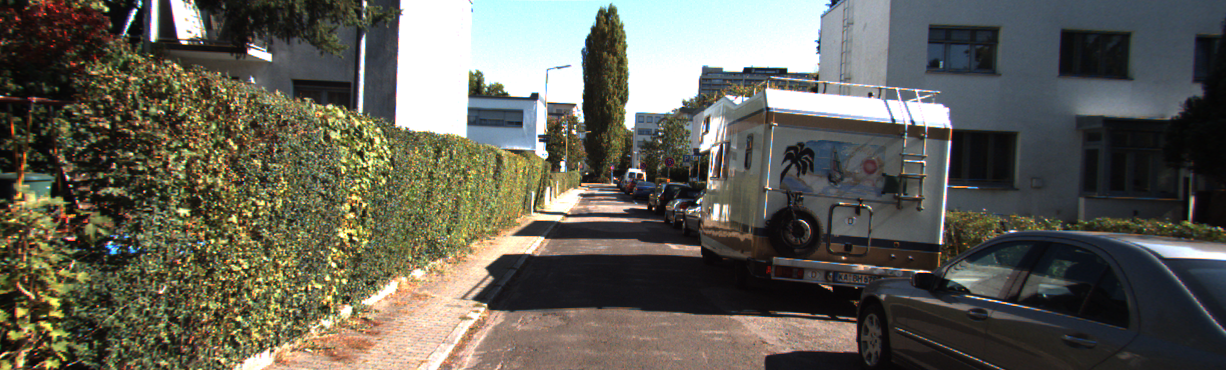}} &
		\includegraphics[width=0.4\columnwidth]{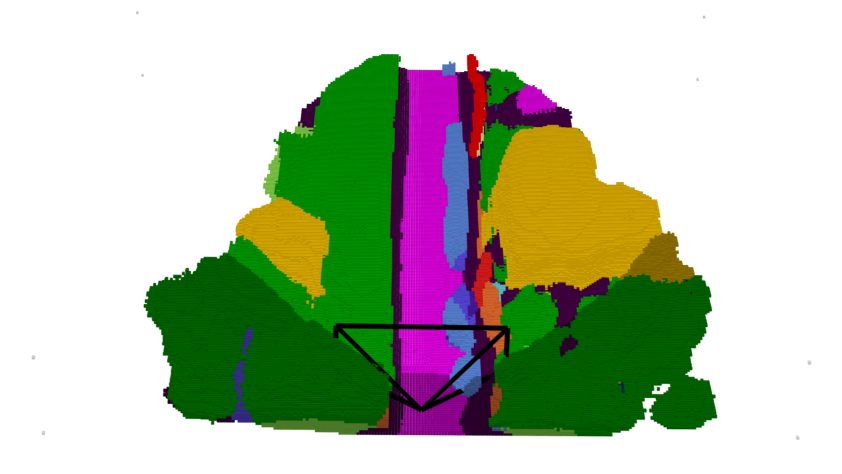} &
		\includegraphics[width=0.4\columnwidth]{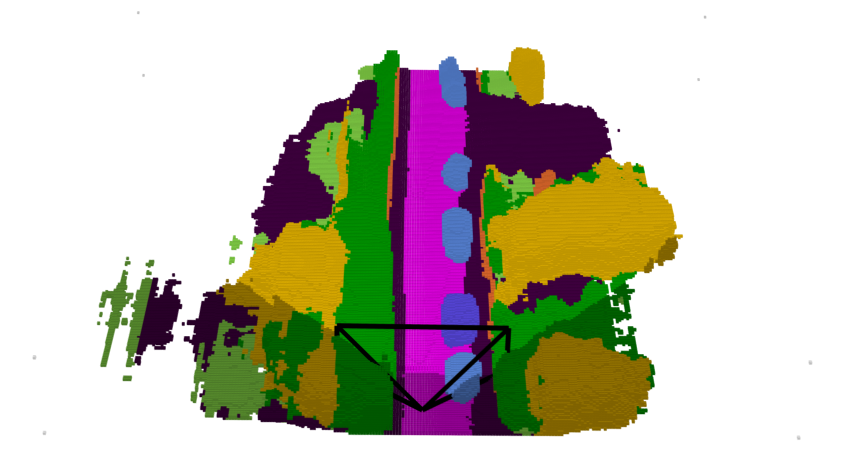} &
		\includegraphics[width=0.4\columnwidth]{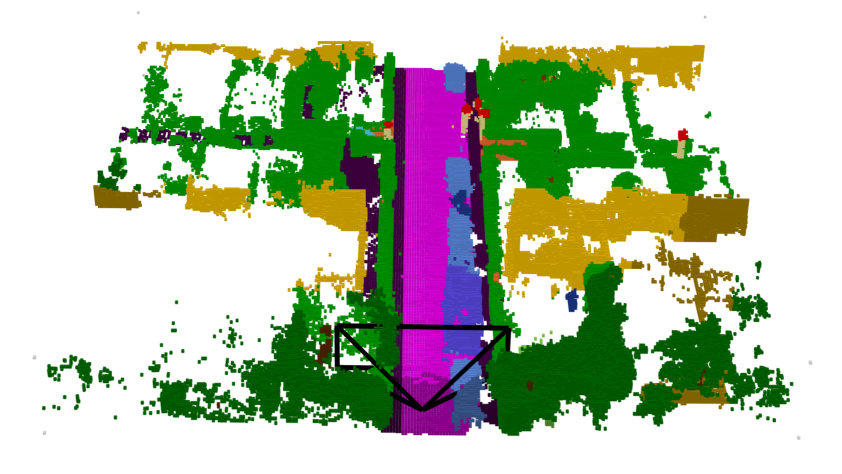} \\
		& \includegraphics[width=0.4\columnwidth]{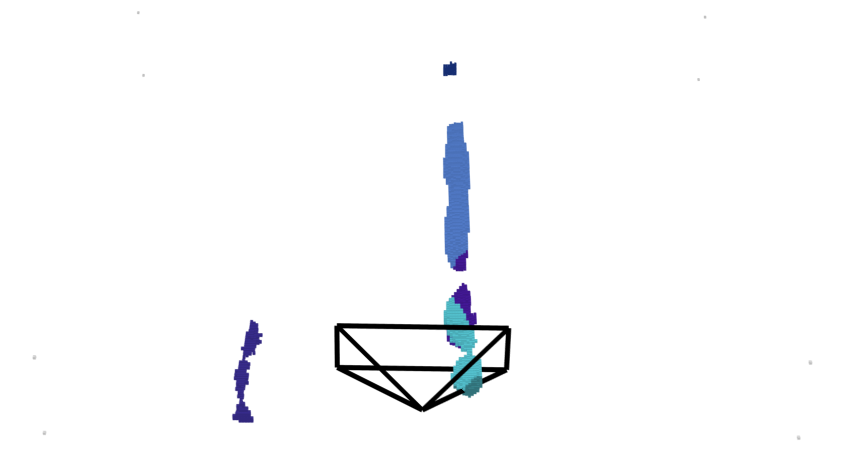} &
		\includegraphics[width=0.4\columnwidth]{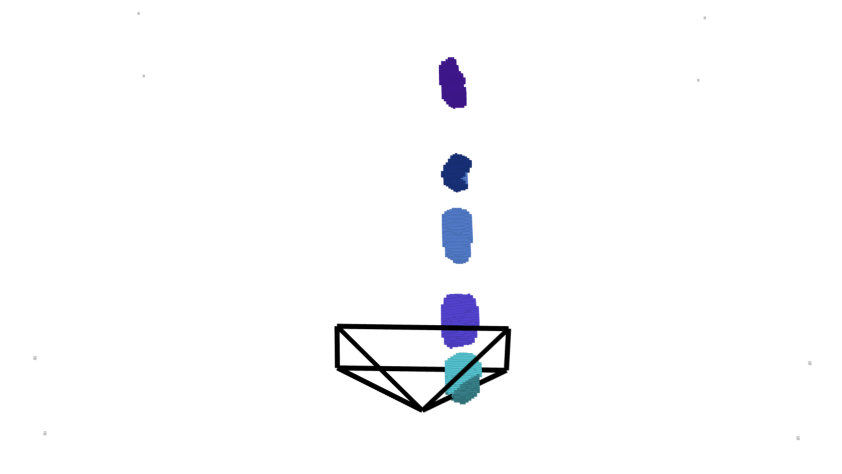} &
		\includegraphics[width=0.4\columnwidth]{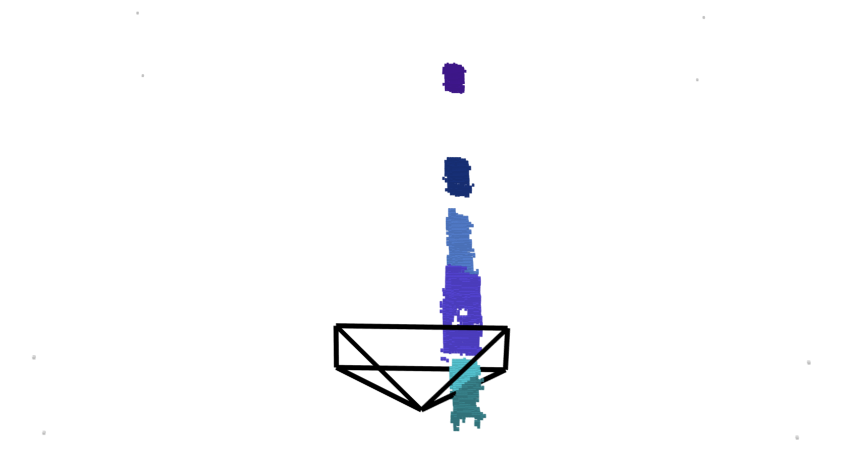} \\
		
		\multirow{2}{*}{\includegraphics[width=0.4\columnwidth]{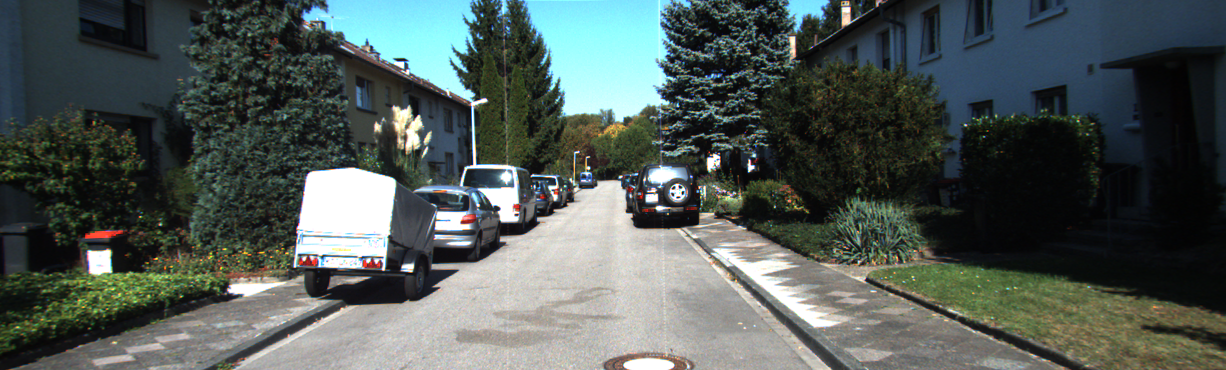}} &
		\includegraphics[width=0.4\columnwidth]{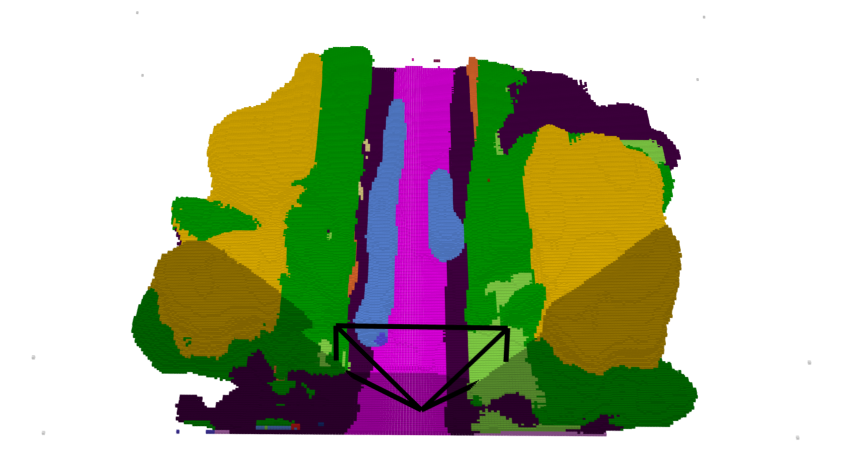} &
		\includegraphics[width=0.4\columnwidth]{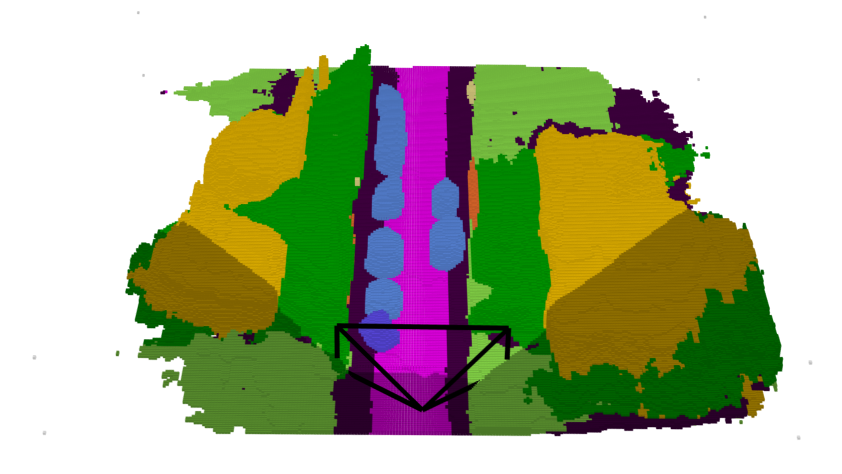} &
		\includegraphics[width=0.4\columnwidth]{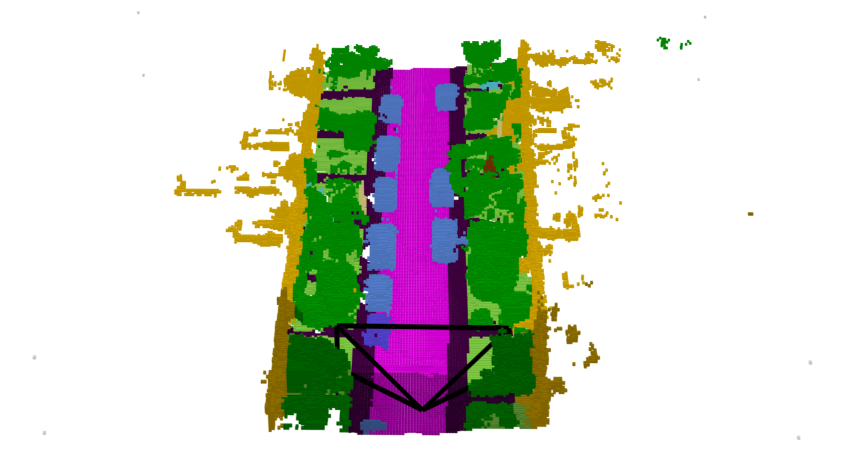} \\
		& \includegraphics[width=0.4\columnwidth]{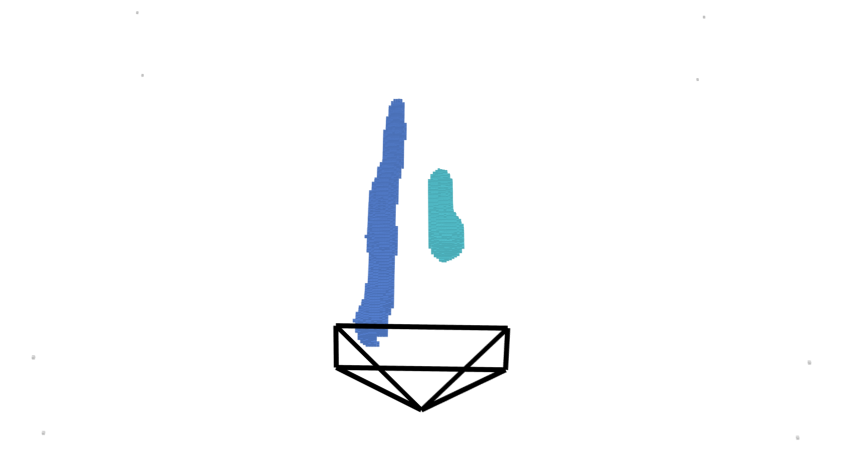} &
		\includegraphics[width=0.4\columnwidth]{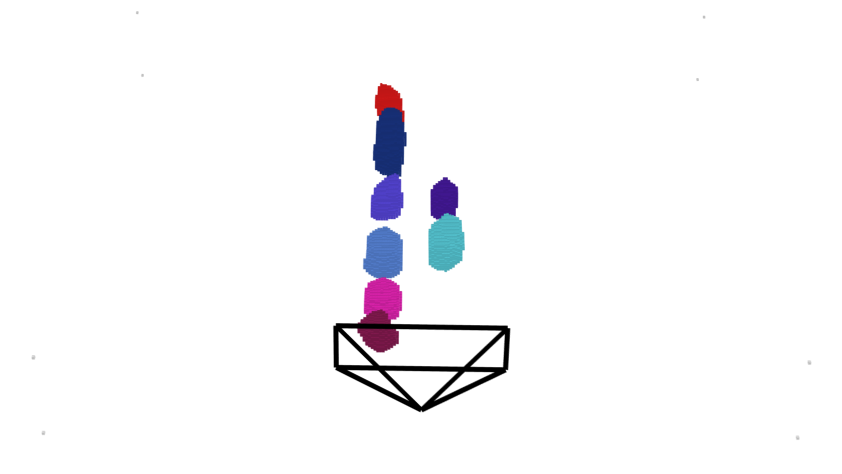} &
		\includegraphics[width=0.4\columnwidth]{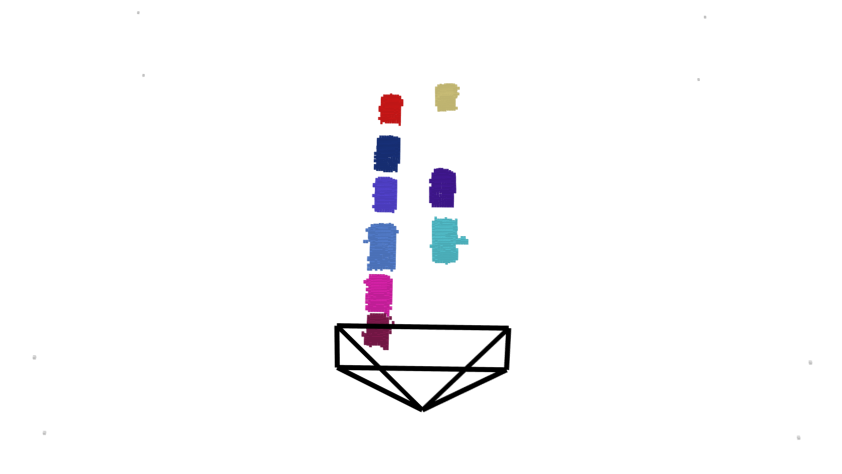} \\
		
		\multirow{2}{*}{\includegraphics[width=0.4\columnwidth]{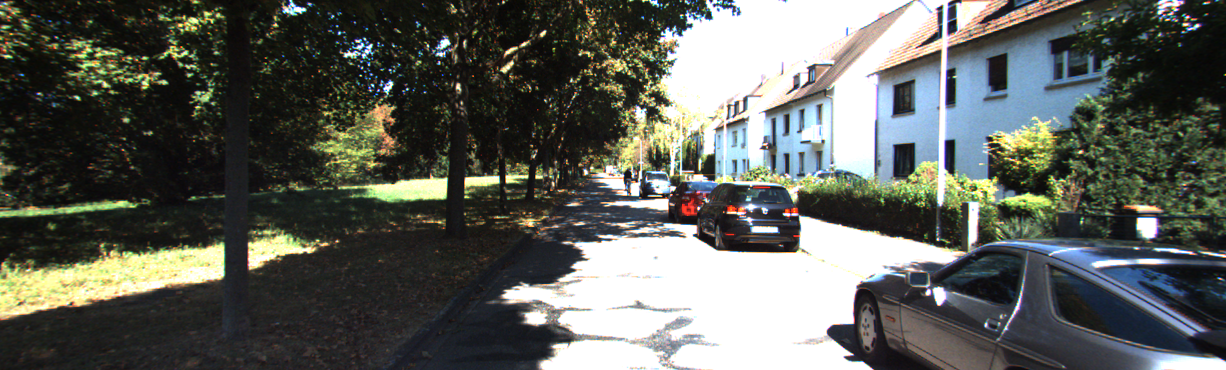}} &
		\includegraphics[width=0.4\columnwidth]{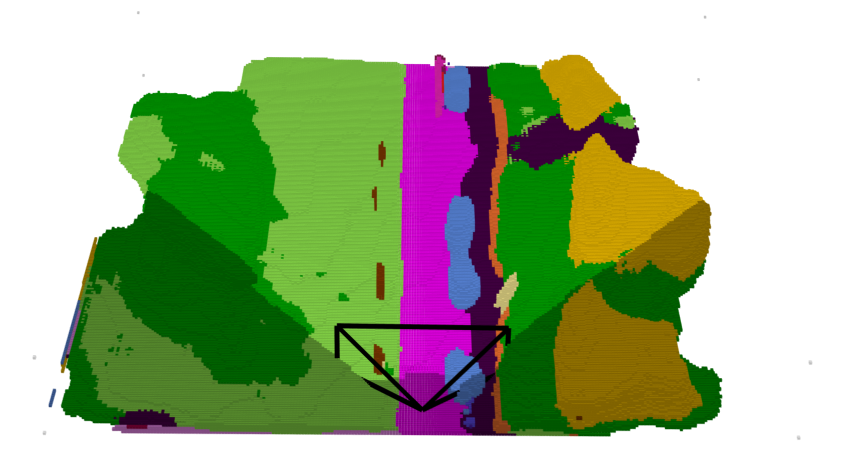} &
		\includegraphics[width=0.4\columnwidth]{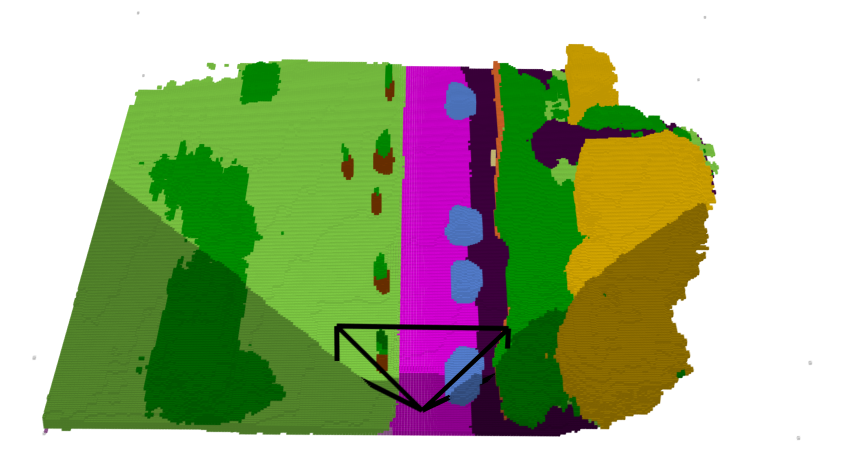} &
		\includegraphics[width=0.4\columnwidth]{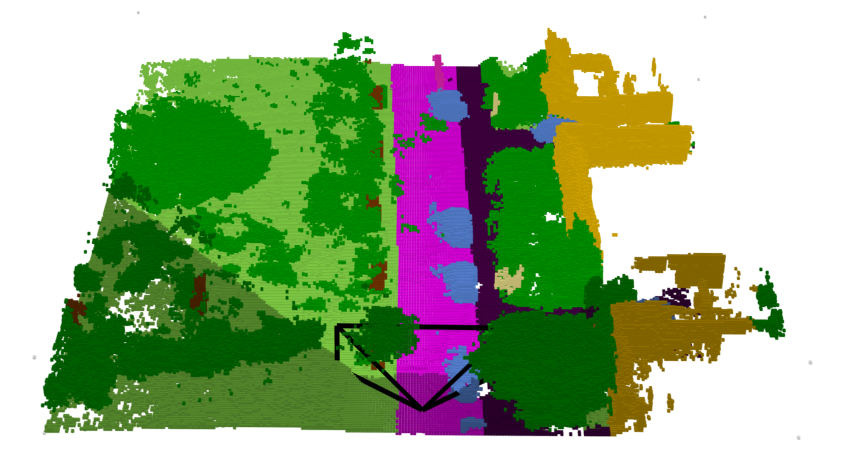} \\
		& \includegraphics[width=0.4\columnwidth]{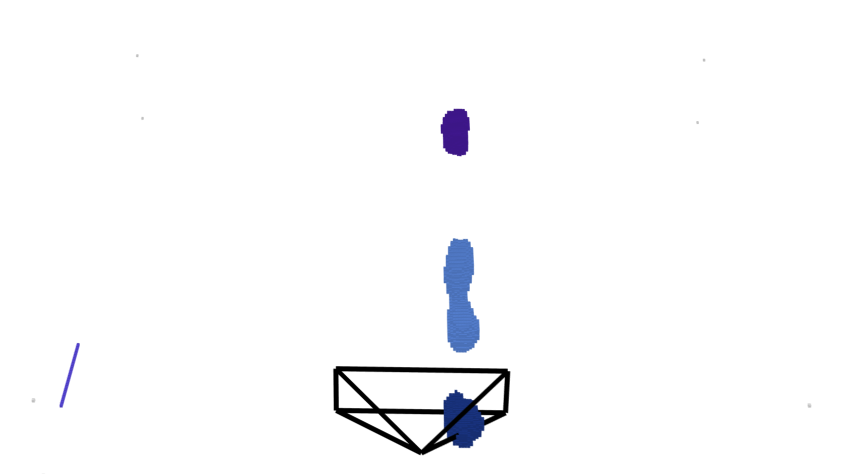} &
		\includegraphics[width=0.4\columnwidth]{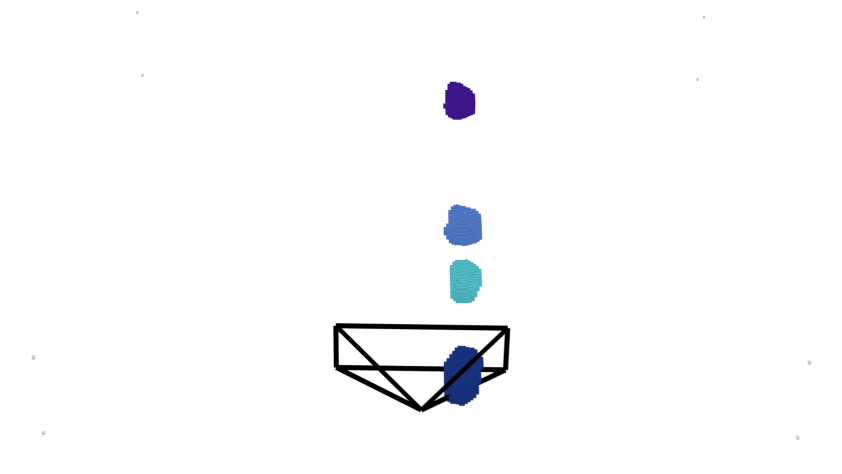} &
		\includegraphics[width=0.4\columnwidth]{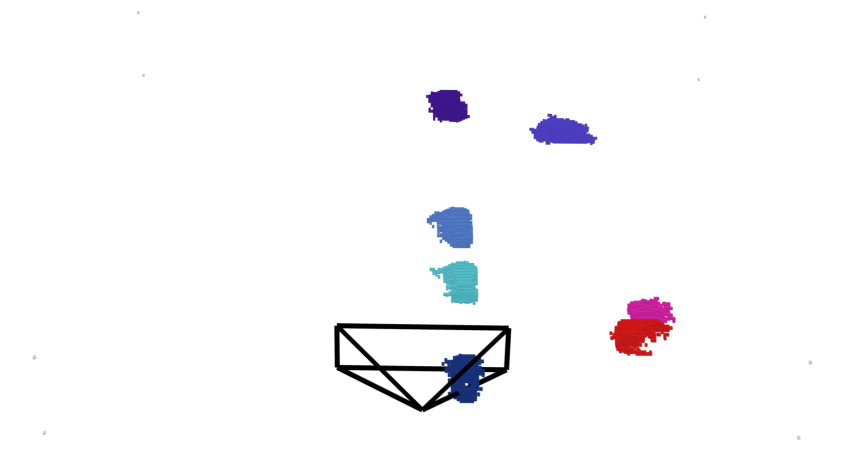} \\
		
		\multirow{2}{*}{\includegraphics[width=0.4\columnwidth]{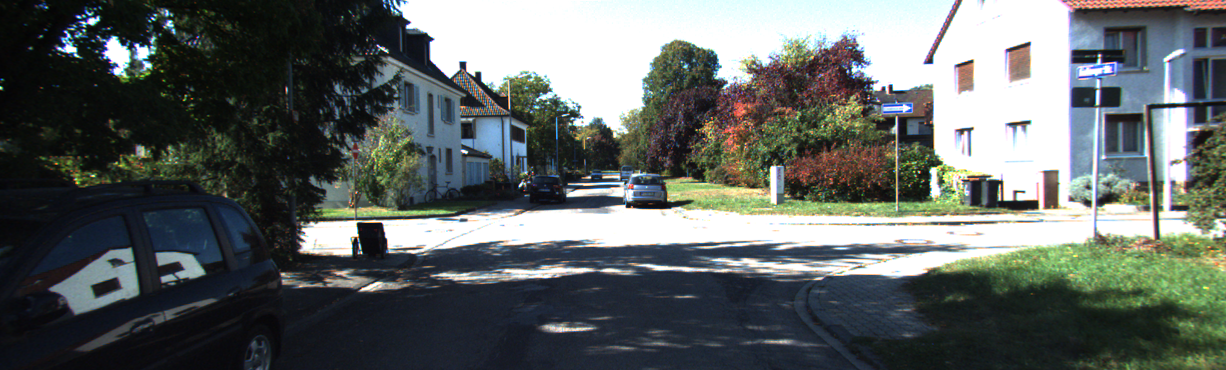}} &
		\includegraphics[width=0.4\columnwidth]{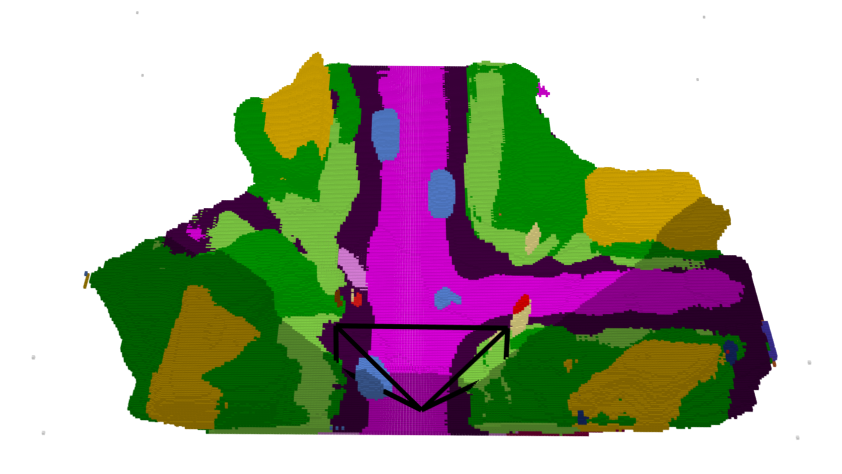} &
		\includegraphics[width=0.4\columnwidth]{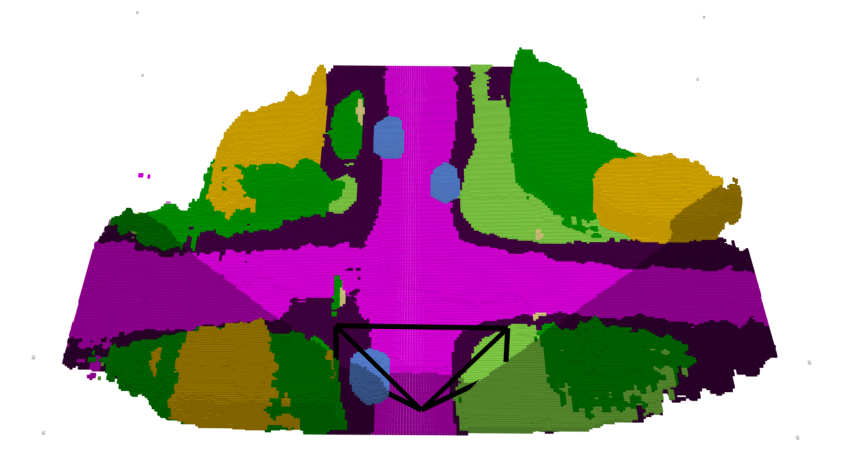} &
		\includegraphics[width=0.4\columnwidth]{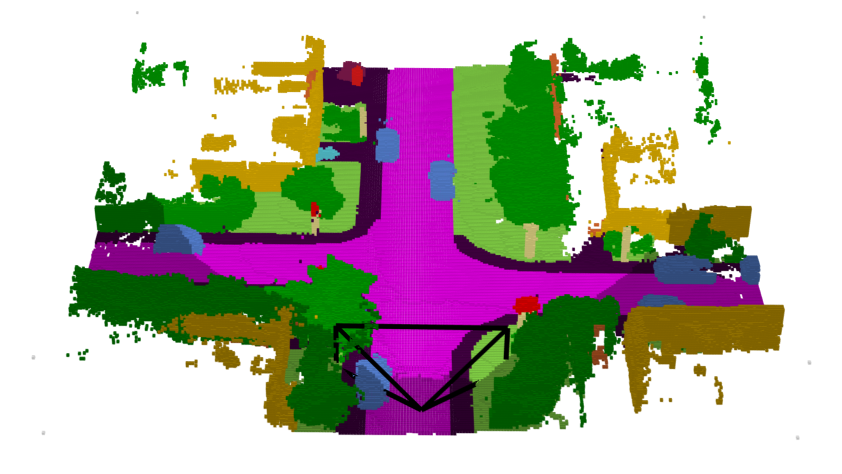} \\
		& \includegraphics[width=0.4\columnwidth]{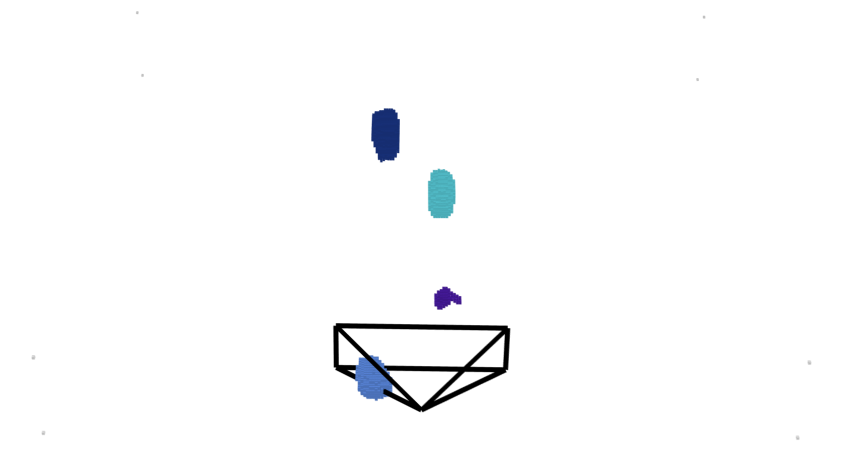} &
		\includegraphics[width=0.4\columnwidth]{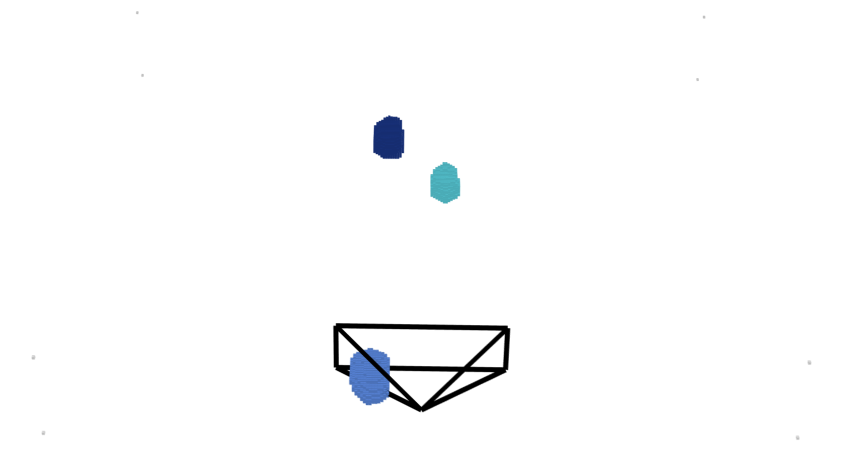} &
		\includegraphics[width=0.4\columnwidth]{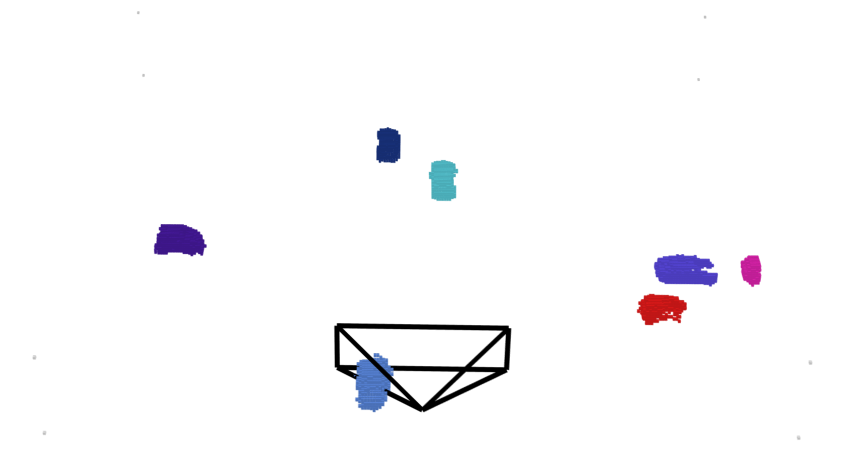} \\
		
		\multicolumn{4}{c}{
		\scriptsize
		\textcolor{bicycle}{$\blacksquare$}bicycle~
		\textcolor{car}{$\blacksquare$}car~
		\textcolor{motorcycle}{$\blacksquare$}motorcycle~
		\textcolor{truck}{$\blacksquare$}truck~
		\textcolor{other-vehicle}{$\blacksquare$}other vehicle~
		\textcolor{person}{$\blacksquare$}person~
		\textcolor{bicyclist}{$\blacksquare$}bicyclist~
		\textcolor{motorcyclist}{$\blacksquare$}motorcyclist~
		\textcolor{road}{$\blacksquare$}road~
		\textcolor{parking}{$\blacksquare$}parking~}\\
		\multicolumn{4}{c}{
		\scriptsize
		\textcolor{sidewalk}{$\blacksquare$}sidewalk~
		\textcolor{other-ground}{$\blacksquare$}other ground~
		\textcolor{building}{$\blacksquare$}building~
		\textcolor{fence}{$\blacksquare$}fence~
		\textcolor{vegetation}{$\blacksquare$}vegetation~
		\textcolor{trunk}{$\blacksquare$}trunk~
		\textcolor{terrain}{$\blacksquare$}terrain~
		\textcolor{pole}{$\blacksquare$}pole~
		\textcolor{traffic-sign}{$\blacksquare$}traffic sign			
		}
	\end{tabular}
	\caption{Visualization on the SemanticKITTI \cite{SemanticKITTI} validation set. Each pair of rows shows the results of semantic scene completion (upper) and 3D instance completion for vehicle (lower). Different color bars represent different categories on the SSC task, while colors indicate different instance for 3D instance completion. The darker voxels are outside FOV of the image. Compared to MonoScene \cite{MonoScene}, our PanoSSC can better capture the road layout (row $7$) and estimate the shape of vehicles (rows $1-6$), especially when they are close. It can also better distinguish similar categories, \eg car and truck (rows $1-4$). }
	\label{fig:qualitative_comparison}
\end{figure*}

\textbf{Metrics.}\label{sec:metrics}
For semantic scene completion, we follow common practices to employ the intersection over union (IoU) of occupied voxels, regardless of their semantic labels, and the mean IoU (mIoU) of 19 semantic classes.

Similar to panoptic 3D scene reconstruction for indoor scenes \cite{panoptic3dss}, we calculate the average of panoptic reconstruction quality (PRQ) of different categories, where $\textrm{PRQ}^c$ for the category $c$ can be written as:
\begin{equation}
	\textrm{PRQ}^c = \frac{\sum_{(h,w,d)\in \textrm{TP}^c}\textrm{IoU}(h,w,d)}{|\textrm{TP}^c|+\frac{1}{2}|\textrm{FP}^c|+\frac{1}{2}|\textrm{FN}^c|},
\end{equation}  
where $\textrm{TP}$, $\textrm{FP}$ and $\textrm{FN}$ are the number of matched pairs of segments, unmatched predicted segments and unmatched ground-truth segments, respectively. Specifically, predicted and ground-truth segments are matched by a greedy search for the maximum IoU, and the match is considered successful if the voxelized IoU $\geq 20\%$. 
% Due to the imperfection of ground truth and the application in autonomous driving, 
We evaluate PRQ of four categories: car, truck, other vehicle and road in SemanticKITTI. For the foreground categories, a segments is the voxels belonging to the same instance id, while all voxels belonging to the road category are a particular background segment. Consistent with the SSC task, we evaluate panoptic reconstruction at a voxel resolution of 0.2m and ignore unknown voxels. In addition, the $\textrm{PRQ}^c$ can be regarded as the product of reconstructed segmentation quality $\textrm{RSQ}^c$ and reconstructed recognition quality $\textrm{RRQ}^c$:
\begin{multline}
	\textrm{PRQ}^c = \textrm{RSQ}^c \times \textrm{RRQ}^c =\\
\frac{\sum_{(h,w,d)\in \textrm{TP}}\textrm{IoU}(h,w,d)}{|\textrm{TP}^c|} \times \frac{|\textrm{TP}^c|}{|\textrm{TP}^c|+\frac{1}{2}|\textrm{FP}^c|+\frac{1}{2}|\textrm{FN}^c|}.
\end{multline}
We also report the average of RSQ and RRQ.

\subsection{Performance}\label{sec:performance}

\textbf{Baselines.} We use the state-of-the-art method MonoScene \cite{MonoScene} as a baseline for semantic scene completion and further cluster the semantic results with Euclidean clustering as the baseline for panoptic 3D scene reconstruction.

% and also report the results of 4 vision-centric baselines in \cite{MonoScene}. For monocular panoptic 3D scene reconstruction, we perform Euclidean clustering consistent with the ground-truth processing on the output of MonoScene\cite{MonoScene} and semantic occupancy head of our network without instance completion head to distinguish different instances. (jk，confusing，ground truth、clustering of mono scence？)
%In the following we detail the performance of PanoSSC on these two tasks.

\textbf{Semantic scene completion.}
\cref{tab:ssc_result} reports the performance of PanoSSC and baselines on SemanticKITTI. 
% As mentioned in \cite{MonoScene}, this is a complex task and the number for all methods is not high. 
Our network achieves performance on par with the state-of-the-art monocular work on the main metric mIoU (11.22 vs 11.27). And the parameter number of PanoSSC is less (137M vs 149M). Besides, our network helps distinguish similar categories and significantly improve the reconstruction of trucks (+5.74) and other vehicles (+3.63). But it is undeniable that PanoSSC's reconstruction of moving objects need to be improved (in SemanticKITTI, for categories like person, there are far more moving objects than stationary ones). We attribute this partly to the imperfection of ground truth in SemanticKITTI mentioned above, that is, moving objects produce traces and do not have the correct shape. Our network performs SSC in the form of reconstructing each instance, which is more susceptible to confusion caused by this imperfection. In addition, PanoSSC infers a global 3D voxel mask for each instance, so the reconstruction accuracy of small object categories also needs improvement.

\begin{table}[htbp]
        \footnotesize
	\centering
	\renewcommand\arraystretch{1.0}

	\begin{tabular}{lcc}
		\toprule
		& mIoU & IoU \\
		\midrule
		Ours w/o instance completion head & 10.59 & 34.95 \\
		Output of semantic occupancy head of ours & 10.77 & \textbf{35.21} \\
		Ours (after merging) & \textbf{11.22} & 34.94 \\
		\bottomrule
	\end{tabular}%
 	\caption{Effect of instance completion head on semantic scene completion in multi-task learning.}
	\label{tab:inst_to_seg}%
\end{table}%

\begin{table}[htbp]
        \footnotesize
	\setlength\tabcolsep{5.0pt}
	\renewcommand\arraystretch{1.0}
	\centering

	\begin{tabular}{lcccc}
		\toprule
		& PRQ   & RSQ   & RRQ   & mIoU \\
		& \multicolumn{4}{c}{things (car,truck,other-vehicle)} \\
		\midrule
		Ours w/o semantic occupancy head  & 5.59  & 21.38 & 28.26 & 6.34 \\
		Ours  & 11.27 & 33.02 & 33.20 & 13.55 \\
		\bottomrule
	\end{tabular}%
 	\caption{Effect of semantic occupancy head on instance completion in multi-task learning.}
	\label{tab:seg_to_inst}%
\end{table}%

\textbf{Panoptic 3D scene reconstruction.}
\cref{tab:panop3dss_result} reports results of panoptic 3D scene reconstruction. Our network evidently outperforms clustering the output of SSC methods. Compared with MonoScene, panoptic reconstruction quality (PRQ) of PanoSSC is higher (+3.60), especially for the foreground categories (+4.76). \cref{tab:panop3dss_result_c} reports the results for each category. Compared with performing Euclidean clustering on the output of semantic occupancy head followed by TPVFormer, adding instance completion head greatly improves PRQ of truck and other-vehicle (+13.26,+1.96). That is, our network can more accurately distinguish these three similar categories: car, truck and other-vehicle.

\textbf{Qualitative results.}
Panoptic 3D scene reconstruction involves semantic completion for background categories and instance completion for foreground categories. \cref{fig:qualitative_comparison} shows the SSC output (upper of each pair of rows) and the instance completion results (lower of each pair of rows). 
%Compared to MonoScene, our network can better estimate the shape of vehicles (rows 1,3,5), especially when they are close. 
MonoScene tends to predict the empty voxels in the vehicle interval as vehicle on the SSC task. Therefore, when simply clustering the output of SSC, it is impossible to assign unique ids to multiple vehicles in the strip (rows 2,4). While PanoSSC can obtain the 3D mask of each instance and merge them, which can better distinguish the close instances and estimate their shape (rows 1,3,5). Besides, since other existing SSC methods use a per-voxel classification formulation, there is a mixture of voxels belonging to the similar categories during semantic occupancy prediction. That is, there are few truck voxels in the region that is mostly predicted to be car voxels (row 1). We adopt mask-wise classification and discard some masks according to overlap conflicts during inference, which can suppress these unreasonable results. In addition, PanoSSC can also better reconstruct road layout (row 7) and distinguish similar categories, \eg car and truck (rows 1,3). Note that none of the existing monocular works can reconstruct the completely occluded objects in the scene well (rows 6,8). More qualitative results are presented in the supplementary material.

\subsection{Ablation studies}\label{sec:ablation}
\textbf{Multi-task learning.}
Inspired by works in 2D domain, our network includes semantic occupancy head and instance completion head for multi-task learning. To prove that these two heads can boost each other, we conduct ablation studies. In \cref{tab:inst_to_seg}, we report the SSC results of our network, the network without instance completion head, and the semantic occupancy head after joint training. It is shown that even without merging the output of instance completion head, joining this head for training can improve SSC performance (mIoU+0.18, IoU+0.26). Merging the output of instance completion head can further boost the main metric of the SSC task (+0.45). \cref{tab:seg_to_inst} shows that ablating semantic occupancy head also impairs the performance of instance completion. We conjecture that this mutual promotion comes from the improvement of generalization by sharing domain information between related tasks.

\begin{table}[tp]
        \footnotesize
	\centering
	\setlength\tabcolsep{5.0pt}
	\renewcommand\arraystretch{1.0}

	\begin{tabular}{lccccc}
		\toprule
		& mIoU & IoU & PRQ   & RSQ   & RRQ \\
		\midrule
		$\lambda_\text{cls}:\lambda_\text{mask}=2:1$   & 10.97 & 34.67 & 21.37 & 39.34 & 44.97 \\
		$\lambda_\text{cls}:\lambda_\text{mask}=1:1$   & 11.03 & 34.78 & 21.54 & 39.51 & 45.00 \\
		$\lambda_\text{cls}:\lambda_\text{mask}=1:2$ (ours) & \textbf{11.22} & \textbf{34.95} & \textbf{22.93} & \textbf{39.51} & \textbf{49.43} \\
		\bottomrule
	\end{tabular}%
 	\caption{Effect of loss weights in instance completion head.}
	\label{tab:loss_weight}%
\end{table}%

\begin{table}[tp]
        \footnotesize
	\centering
	\setlength\tabcolsep{6.0pt}
	\renewcommand\arraystretch{1.0}

	\begin{tabular}{lccc}
		\toprule
		Layer & PRQ   & RSQ   & RRQ \\
		\midrule
		1     & 21.13 & 39.44 & 44.18 \\
		2     & 21.78 & 38.97 & 46.38 \\
		3     & 22.93 & 39.52 & 49.43 \\
		\bottomrule
	\end{tabular}%
 	\caption{Panoptic 3D scene reconstruction results for each layer in instance completion head.}
	\label{tab:maskdecoderlayer}%
\end{table}%

\textbf{Losses.}
The loss of instance completion head in \cref{eq:maskloss} consists of classification loss and mask loss, and we need to balance these two losses. We find that classification loss converges slightly faster than mask loss. \cref{tab:loss_weight} shows that appropriately reducing the weight of classification loss is beneficial to the network to obtain good results. We speculate that this is because classification is easier than estimation of shape and position for mask decoder.

\textbf{Mask decoder.}
As mentioned in \cref{sec:maskdecoder}, instance completion head is stacked by multiple transformer layers and each layer can generate a set of classification probabilities and 3D masks. \cref{tab:maskdecoderlayer} reports the results of each layer in instance completion head. As the number of layers increases, the reconstruction quality of the output improves. Considering the parameter amount and inference speed of the network, our PanoSSC only stacks 3 layers.

\section{Conclusion}\label{sec:conclu}
In this paper, we proposed a novel voxelized scene understanding method, coined PanoSSC, which can tackle semantic occupancy prediction and panoptic 3D scene reconstruction on outdoor. Our method joins semantic occupancy head and instance completion head for joint training to achieve mutual promotion. On the SemanticKITTI dataset, we perform on par with the state-of-the-art monocular method on semantic occupancy prediction task. And to our best knowledge, PanoSSC is the first vision-only panoptic 3D scene reconstruction method on outdoor and achieves good results. We hope that our work can advance the research on more comprehensive scene understanding for autonomous driving.

{
    \small
    \bibliographystyle{ieeenat_fullname}
    \bibliography{ref/ssc}
}
\clearpage
\setcounter{page}{1}
\maketitlesupplementary 
\section{3D mask decoder}
As described in \cite{PanopticSegformer}, using a lightweight FC layer to generate masks from attention maps enables the attention module to learn where to focus guided by the ground truth mask. We extend the mask decoder in \cite{PanopticSegformer} to 3D segmentation and completion, and also use deep supervision, which means that attention maps of each layer are supervised by the ground truth 3D mask. Therefore, the attention module can focus on interested region as early as possible, which accelerating the learning and convergence of the model.

\section{Dataset setup.}
\begin{figure}[htbp]
	\centering
	\includegraphics[width=0.47\textwidth]{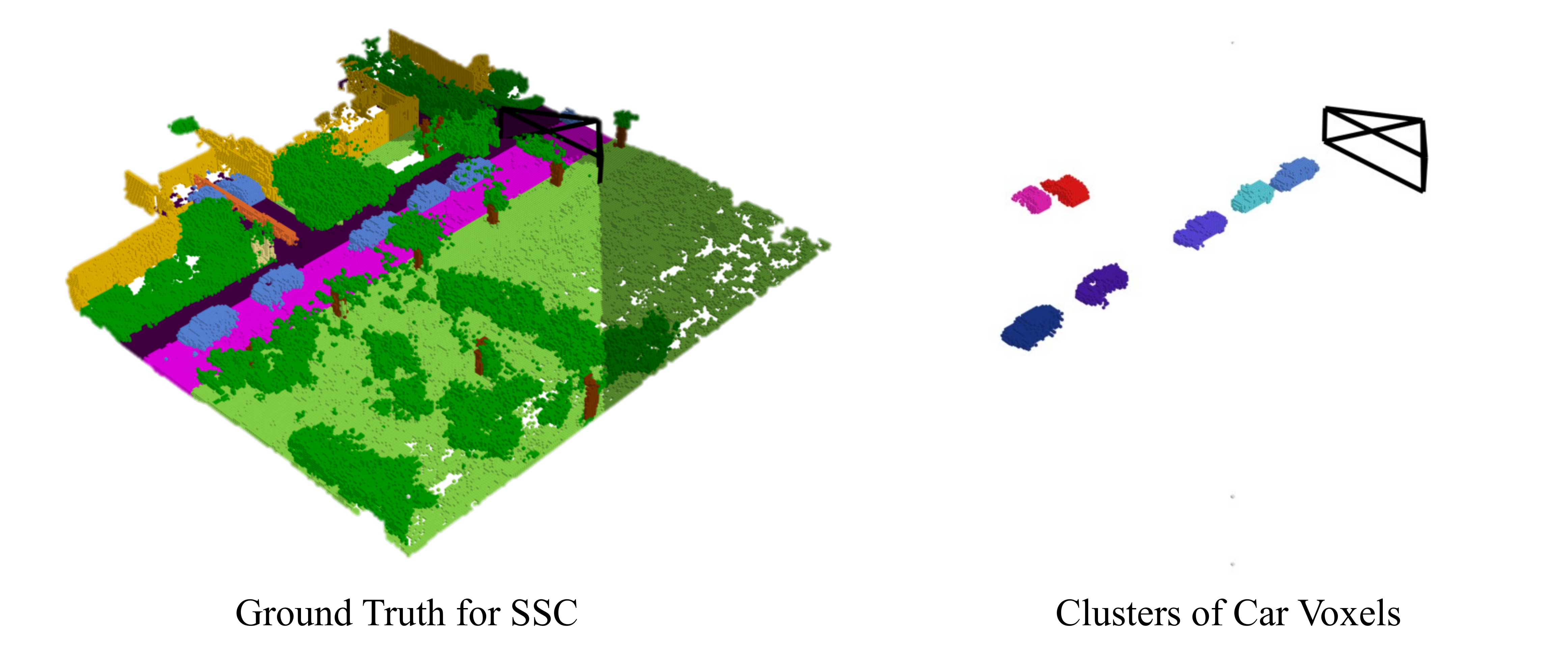}
	%\vspace{-7mm}
	\caption{Instance ids obtained from Euclidean clustering on the ground truth for SSC.
	}
	\label{fig:cluster}
	%\vspace{-6mm}
\end{figure}

When training instance completion head, the 3D binary mask and the category label for each instance is required. While SemanticKITTI only provides ground-truth semantic labels without instance ids and moving objects inevitably produce traces. To train and evaluate our network, we perform Euclidean clustering on the ground truth and filter out those traces. In practice, we set the search radius of Euclidean clustering to be 2 voxels for vehicles and 3 voxels for other categories. The maximum number of voxels per cluster is 2000 for cars, 5000 for trucks and other-vehicles, and 1000 for people, bicyclists and motorcyclists. As shown in \cref{fig:cluster}, we can obtain unique ids for different instances by setting the reasonable clustering parameters.

\section{Qualitative results}\label{sec:more_qualitative}
\begin{figure*}
	\centering
	\scriptsize
	\begin{tabular}{cccc}
		Input & MonoScene \cite{MonoScene} & PanoSSC (ours) & Ground Truth\\
		\multirow{2}{*}{\includegraphics[width=0.4\columnwidth]{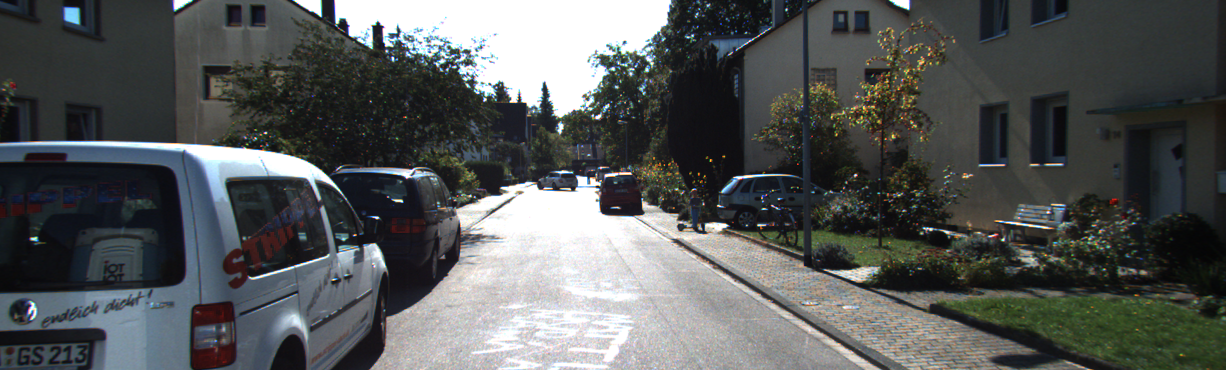}} &
		\includegraphics[width=0.4\columnwidth]{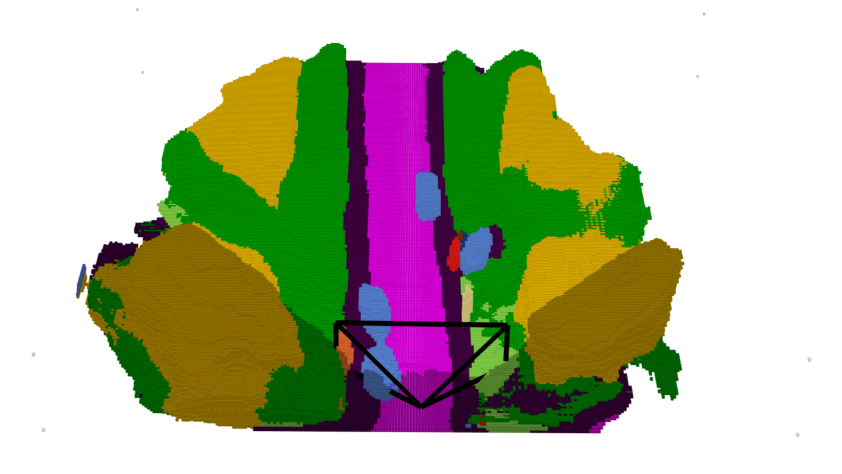} &
		\includegraphics[width=0.4\columnwidth]{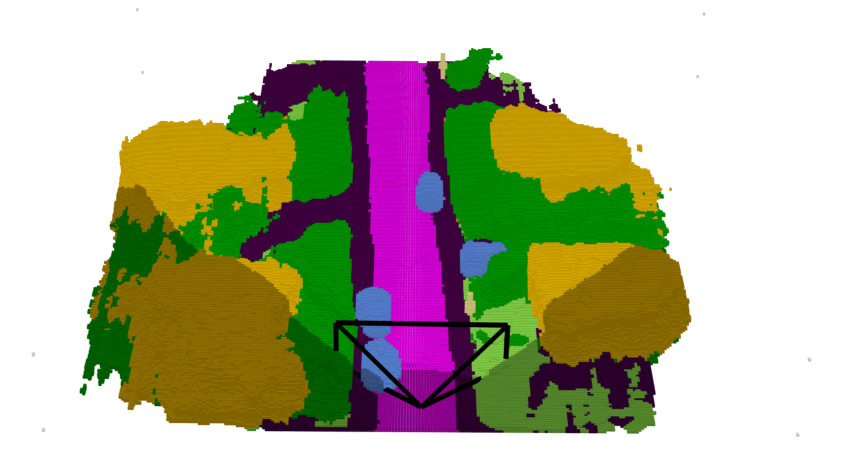} &
		\includegraphics[width=0.4\columnwidth]{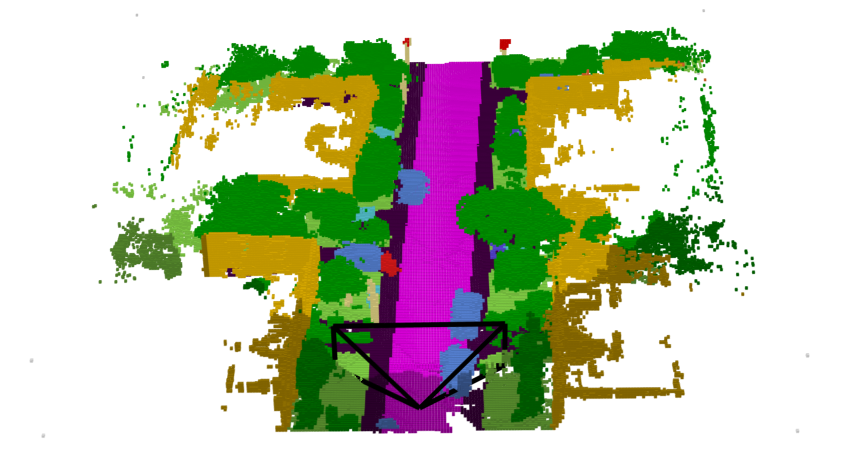} \\
		& \includegraphics[width=0.4\columnwidth]{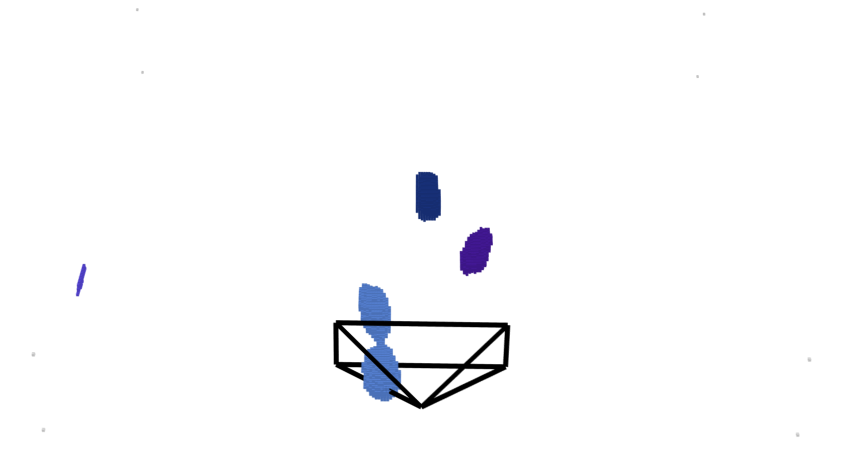} &
		\includegraphics[width=0.4\columnwidth]{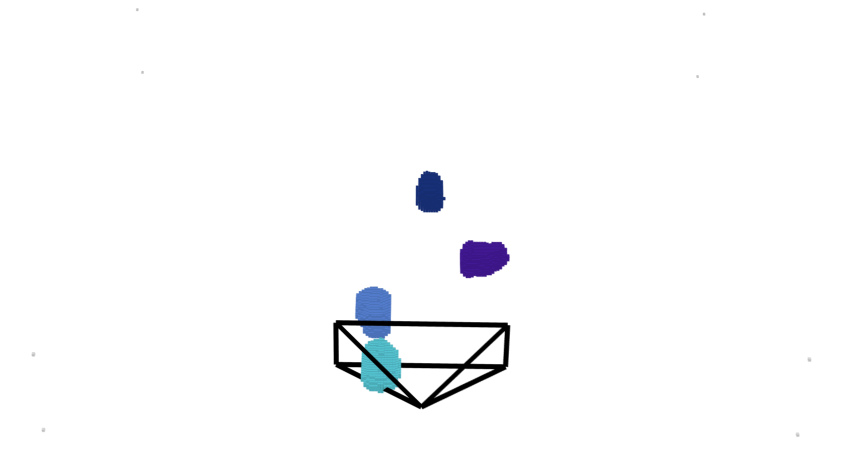} &
		\includegraphics[width=0.4\columnwidth]{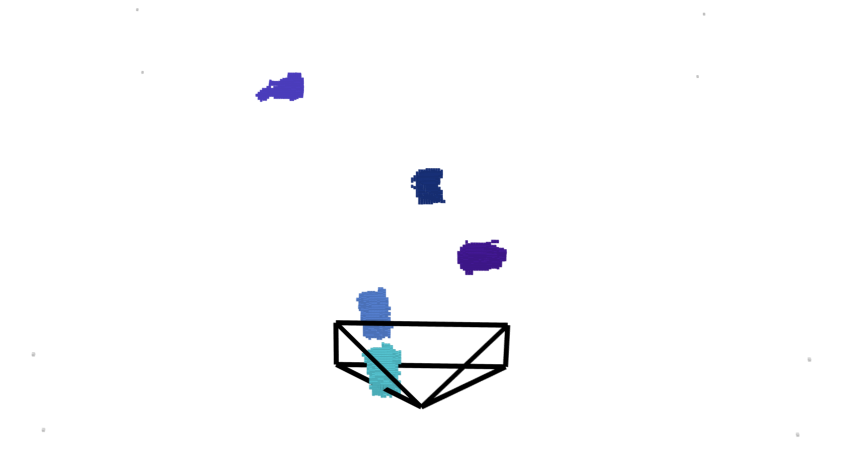} \\
		
		\multirow{2}{*}{\includegraphics[width=0.4\columnwidth]{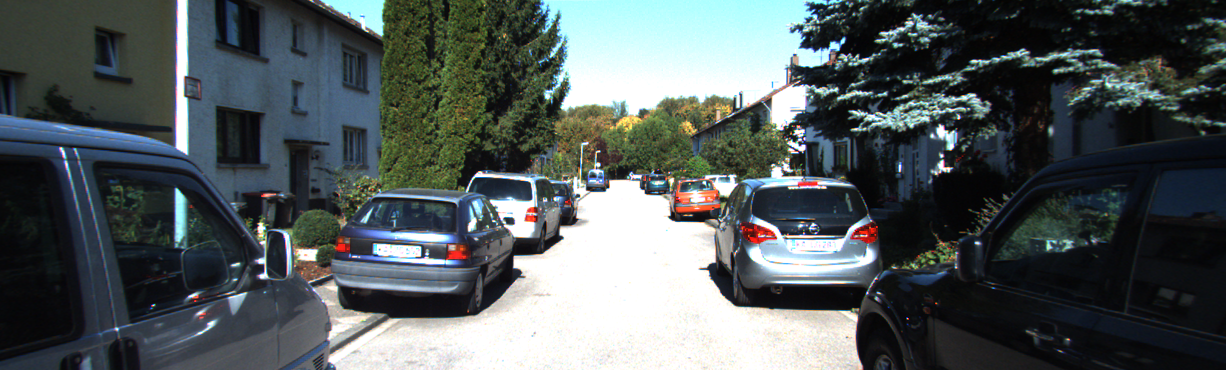}} &
		\includegraphics[width=0.4\columnwidth]{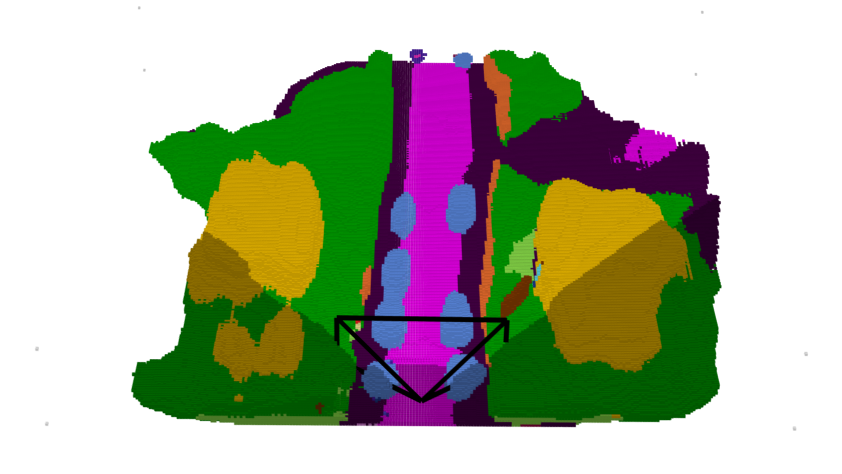} &
		\includegraphics[width=0.4\columnwidth]{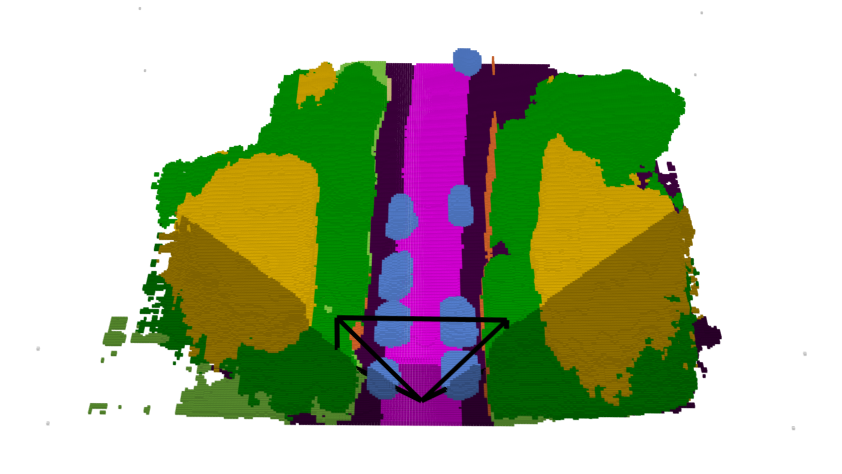} &
		\includegraphics[width=0.4\columnwidth]{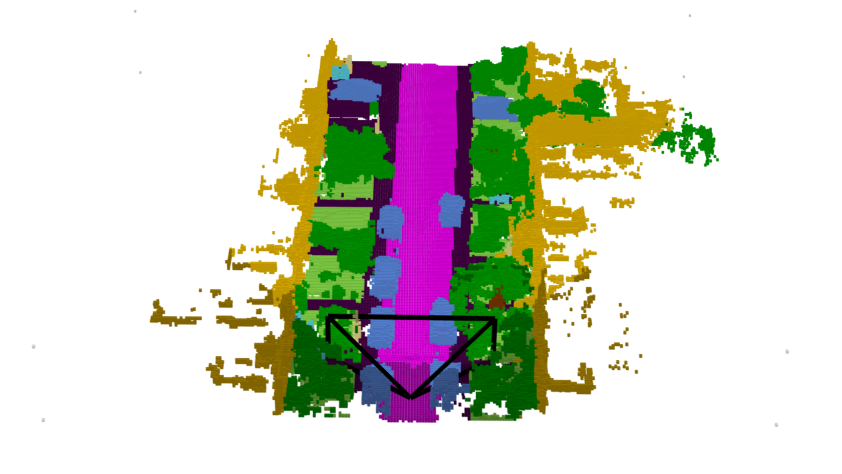} \\
		& \includegraphics[width=0.4\columnwidth]{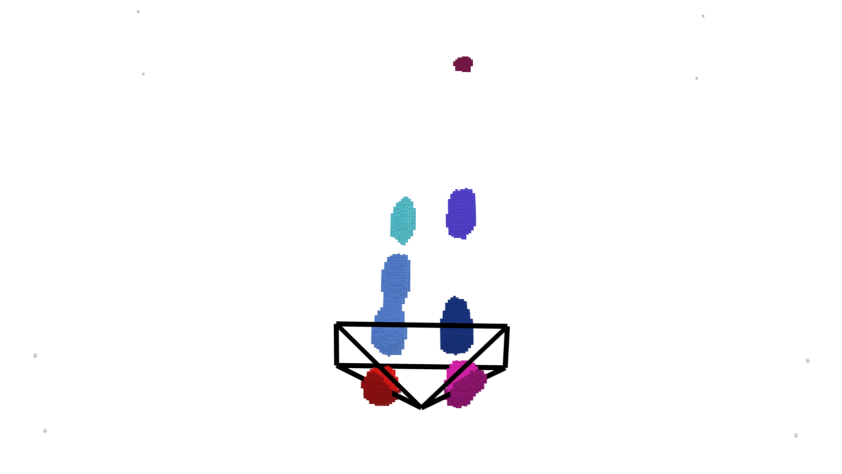} &
		\includegraphics[width=0.4\columnwidth]{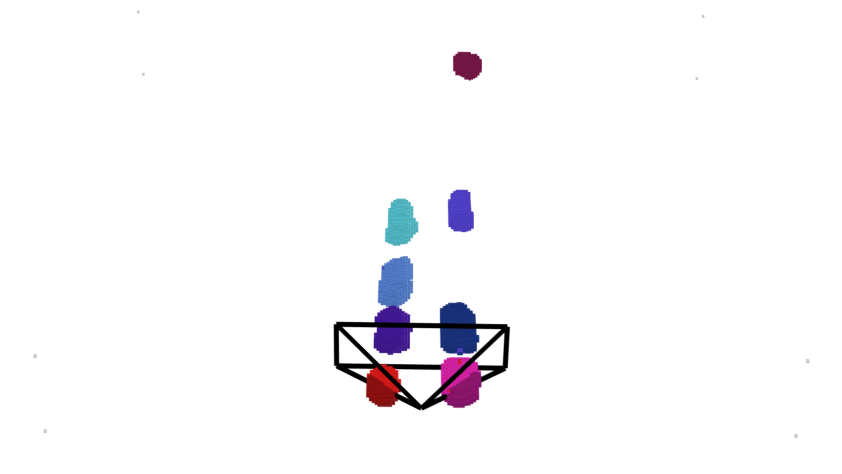} &
		\includegraphics[width=0.4\columnwidth]{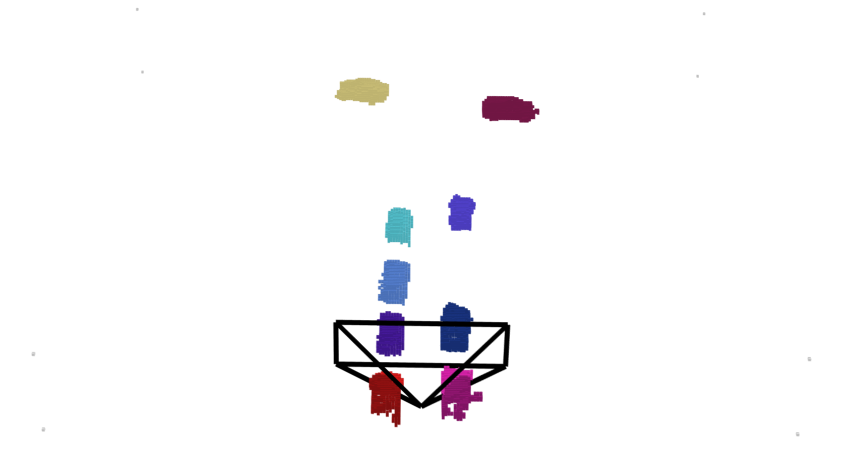} \\
		
		\multirow{2}{*}{\includegraphics[width=0.4\columnwidth]{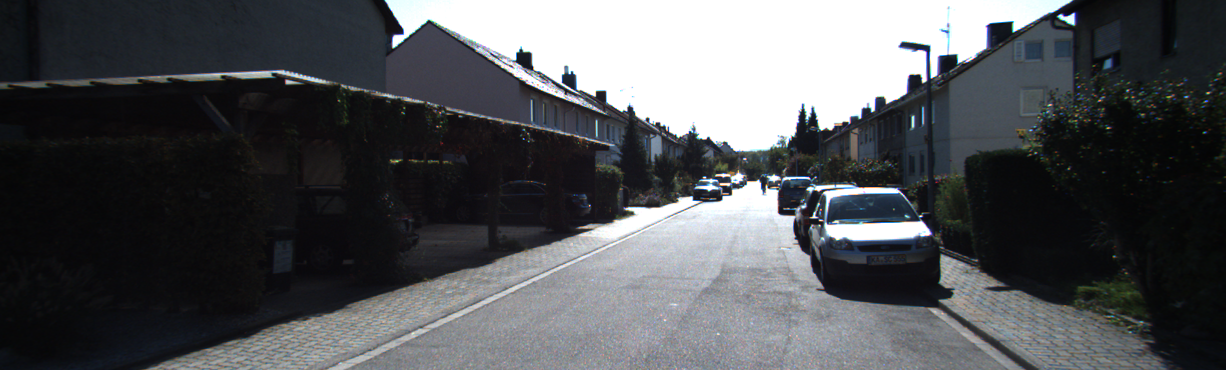}} &
		\includegraphics[width=0.4\columnwidth]{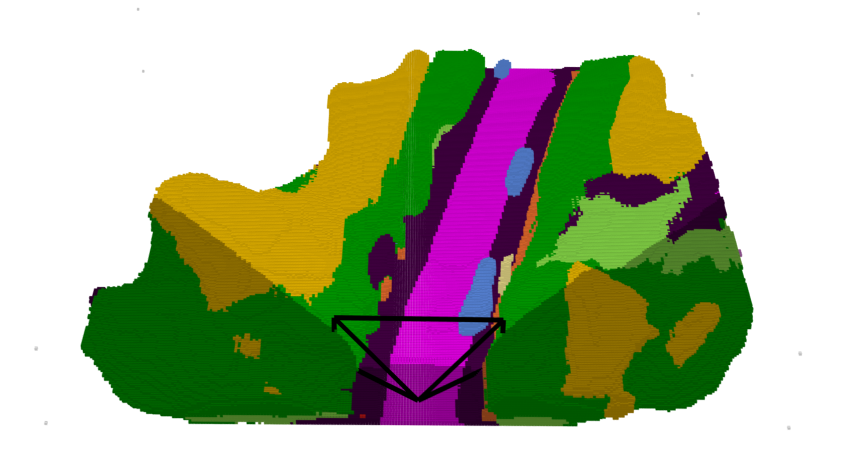} &
		\includegraphics[width=0.4\columnwidth]{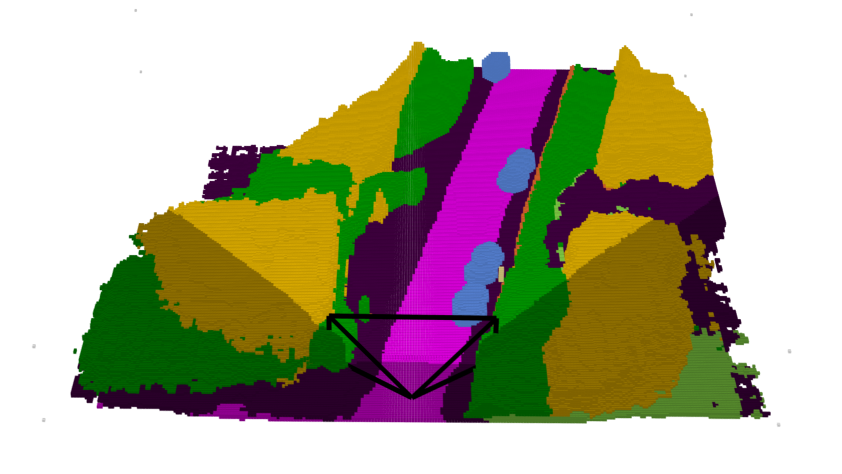} &
		\includegraphics[width=0.4\columnwidth]{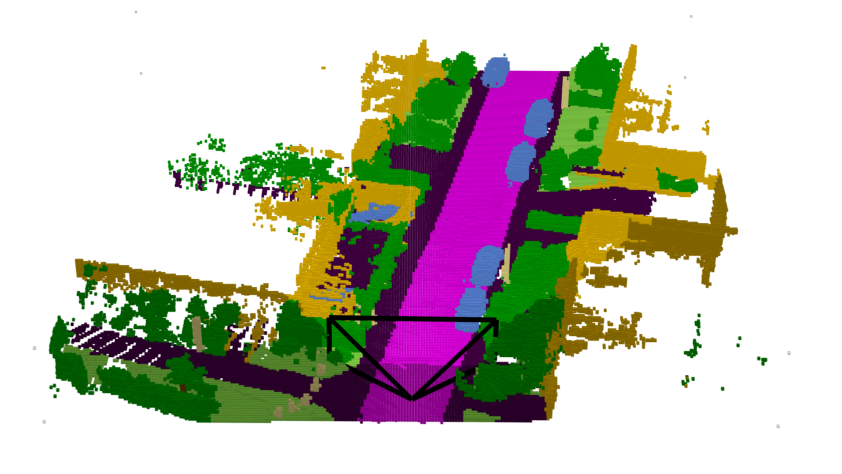} \\
		& \includegraphics[width=0.4\columnwidth]{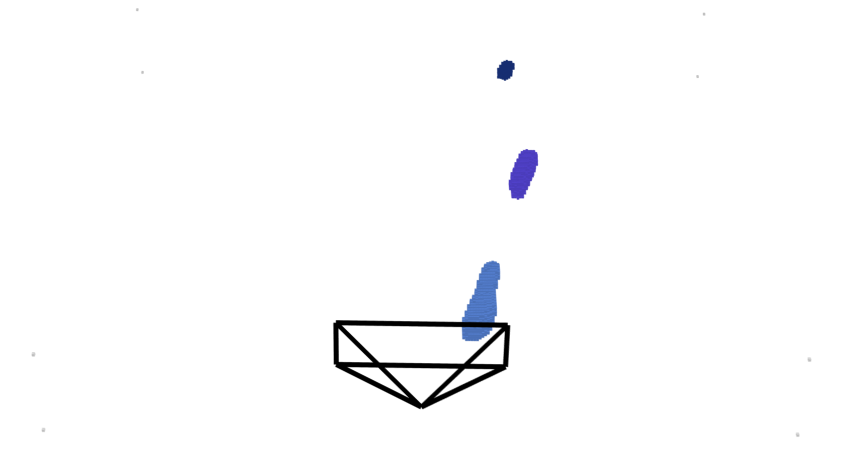} &
		\includegraphics[width=0.4\columnwidth]{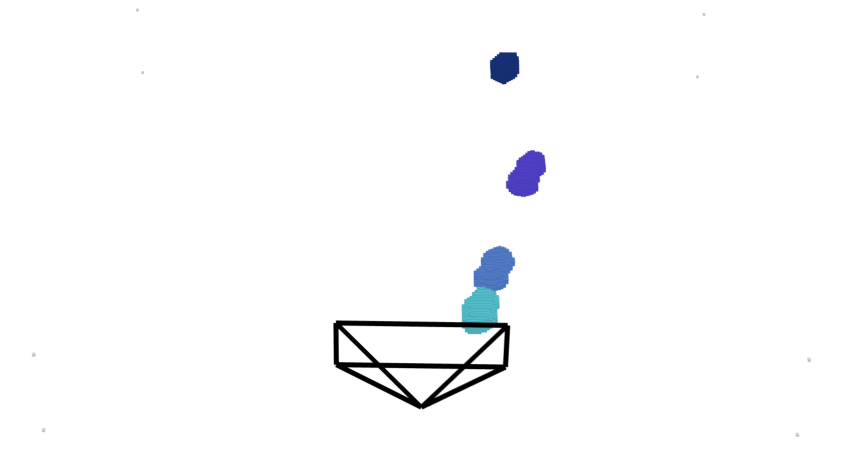} &
		\includegraphics[width=0.4\columnwidth]{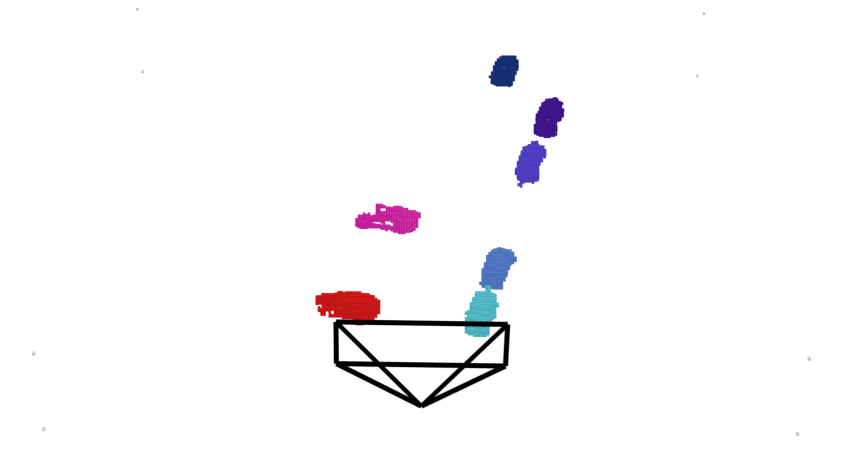} \\

		\multicolumn{4}{c}{
			\scriptsize
			\textcolor{bicycle}{$\blacksquare$}bicycle~
			\textcolor{car}{$\blacksquare$}car~
			\textcolor{motorcycle}{$\blacksquare$}motorcycle~
			\textcolor{truck}{$\blacksquare$}truck~
			\textcolor{other-vehicle}{$\blacksquare$}other vehicle~
			\textcolor{person}{$\blacksquare$}person~
			\textcolor{bicyclist}{$\blacksquare$}bicyclist~
			\textcolor{motorcyclist}{$\blacksquare$}motorcyclist~
			\textcolor{road}{$\blacksquare$}road~
			\textcolor{parking}{$\blacksquare$}parking~}\\
		\multicolumn{4}{c}{
			\scriptsize
			\textcolor{sidewalk}{$\blacksquare$}sidewalk~
			\textcolor{other-ground}{$\blacksquare$}other ground~
			\textcolor{building}{$\blacksquare$}building~
			\textcolor{fence}{$\blacksquare$}fence~
			\textcolor{vegetation}{$\blacksquare$}vegetation~
			\textcolor{trunk}{$\blacksquare$}trunk~
			\textcolor{terrain}{$\blacksquare$}terrain~
			\textcolor{pole}{$\blacksquare$}pole~
			\textcolor{traffic-sign}{$\blacksquare$}traffic sign			
		}
	\end{tabular}
	\caption{Additional qualitative results on the SemanticKITTI \cite{SemanticKITTI} validation set. Each pair of rows shows the results of semantic scene completion (upper) and 3D instance completion for vehicle (lower). Different color bars represent different categories in SSC task, while colors indicate different instance for 3D instance completion. The darker voxels are outside FOV of the image.}
	\label{fig:additional_qualitative}
\end{figure*}

\begin{figure*}
	\centering
	\scriptsize
	\begin{tabular}{ccc}
		Input & MonoScene \cite{MonoScene} & PanoSSC (ours) \\
		\includegraphics[width=0.8\columnwidth]{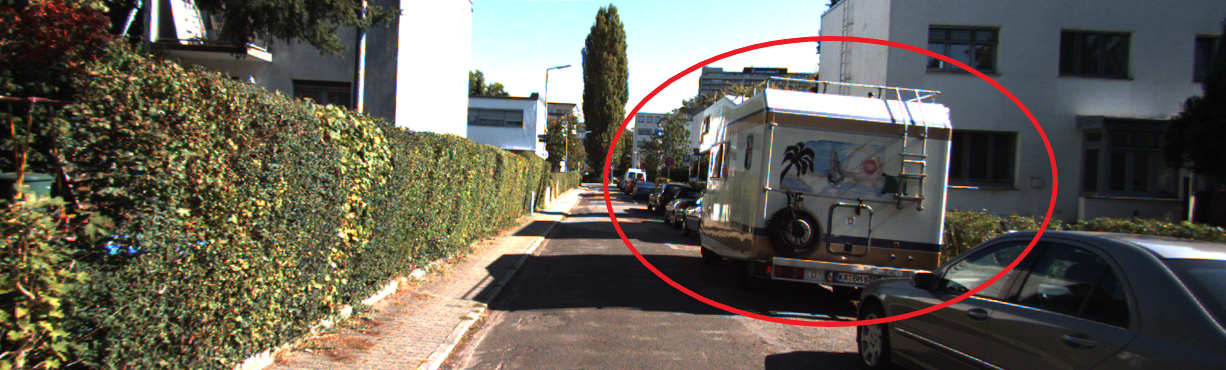} &
		\includegraphics[width=0.4\columnwidth]{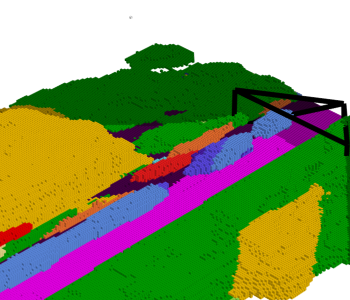} &
	    \includegraphics[width=0.4\columnwidth]{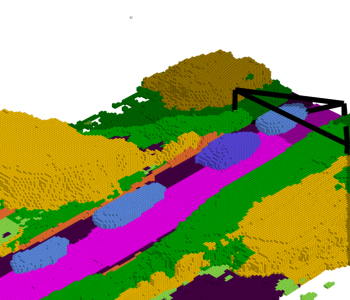} \\
				
		\multicolumn{3}{c}{
			\scriptsize
			\textcolor{car}{$\blacksquare$}car~
			\textcolor{truck}{$\blacksquare$}truck~
			\textcolor{road}{$\blacksquare$}road~
			\textcolor{sidewalk}{$\blacksquare$}sidewalk~
			\textcolor{building}{$\blacksquare$}building~
			\textcolor{vegetation}{$\blacksquare$}vegetation~
			\textcolor{terrain}{$\blacksquare$}terrain~
			\textcolor{pole}{$\blacksquare$}pole~
			\textcolor{traffic-sign}{$\blacksquare$}traffic sign}\\
	\end{tabular}
	\caption{Zoom-in view of a mixture of voxels belonging to the similar categories during semantic occupancy prediction.}
	\label{fig:mixture}
\end{figure*}

\cref{fig:additional_qualitative} shows additional qualitative results on SemanticKITTI validation set. Notice our network can better distinguish the close instances and estimate their shape. \cref{fig:mixture} provides zoom-in view of a mixture of voxels belonging to the car and truck categories. It can be clearly shown that compared with the per-voxel classification commonly used in SSC methods, the mask-wise classification and merging strategy in PanoSSC can suppress these unreasonable results.

\end{document}